%% file: main.tex
\documentclass[11pt]{article}

\usepackage[margin=1in]{geometry}
\usepackage{amsmath, amssymb, amsfonts, mathtools, amsthm}
\usepackage{booktabs}
\usepackage{array}
\usepackage{enumitem}
\usepackage{float}
\usepackage{graphicx}
\usepackage[authoryear]{natbib}
\usepackage{xcolor}
\PassOptionsToPackage{hyphens}{url}
\usepackage[colorlinks=true, allcolors=blue!60!black]{hyperref}

\input{macros}

\title{Prediction-Powered Smoothing and Validation for Disaggregated AI Evaluation}
\author{ %
  Sho Kawano\thanks{Corresponding author: \href{mailto:shkawano@ucsc.edu}{shkawano@ucsc.edu}}
  \quad Zehang Richard Li \quad Paul A. Parker  \\[0.4em]
  \normalsize Department of Statistics, University of California, Santa Cruz, CA, USA
}
\date{\today}

\begin{document}
\maketitle

\begin{abstract}
Evaluating an AI system requires disaggregated assessment, as performance varies across domains such as benchmark task types or conversation types in deployed agents.
Exhaustive testing is expensive, so evaluation rests on a sample of labeled units.
We treat the evaluation set as a finite population and seek accurate point and interval estimates of each domain mean.
Direct estimators, including prediction-powered inference (PPI), use only a domain's own labels and are imprecise where labels are few.
Small area estimation addresses this problem, and we build on it to develop an integrated workflow for estimation and validation.
For estimation, we propose prediction-powered smoothing (PP-S), a Bayesian model fit to each domain's prediction-powered estimate, with an extension that borrows strength across a reporting taxonomy (PP-TS).
For validation, we derive a new, approximately unbiased design-based cross-validation score for choosing among direct and smoothed estimators.
We study a curated benchmark with verifiable grading and deployed agent traffic graded by humans, each with every outcome observed.
In both, the proposed estimators improve on the direct estimators in point and interval estimation, with near-nominal coverage.
At the same sampling budget, our score selects as well as an independent validation sample does and estimates the selected estimator's error far more accurately.
\end{abstract}

{\raggedright\noindent\textbf{Keywords:} disaggregated evaluation; prediction-powered inference; small area estimation; cross-validation; survey sampling\par}

\section{Introduction}\label{sec:intro}

AI systems built on large language models (LLMs) are rapidly being deployed in a wide range of settings, increasing the number of systems and versions to evaluate. 
A single headline score can mask heterogeneous performance. 
Evaluators therefore report performance separately for each task domain, product line or user segment the system serves.
This practice is called disaggregated evaluation \citep{barocas2021}.

Disaggregated evaluation can be expensive because observing an outcome may require querying the system, grading its response, or both.
An evaluation unit could be a single task in an LLM benchmark or a single customer-support conversation with a deployed system.
Reporting dimensions (e.g., task category) partition the population into domains.
The outcome for each unit is a grade for how well the task was completed or a binary indicator of successful completion.
The target is the mean outcome over all units in a given domain.
Obtaining an outcome takes two steps: querying, in which the system carries out the task; and grading, in which the attempt is scored. 
We distinguish two settings by the grading step.
\begin{itemize}
\item \textbf{Verifiable.} Each task has a known answer, so grading can be done programmatically.
\item \textbf{Open-ended.} Correctness cannot be checked against a known answer, so the gold standard often needs to be produced by a human grader.
\end{itemize}
For verifiable tasks, grading is automatic and the cost lies in querying.
Evaluating a single LLM on HELM, a major benchmark suite, can cost \$9{,}337 in API credits \citep{liang2022helm}.
For open-ended tasks, human assessment creates an additional cost.
In monitoring the traffic of a deployed AI system, the interactions between human and AI have already occurred, so the remaining cost is only in reviewing and grading them.
A benchmark of open-ended tasks, such as replicating a research paper \citep{starace2025} or answering a medical question at length \citep{singhal2023}, requires both token-intensive querying and expert grading.
As the number of units, systems, and model versions grows, exhaustive evaluation becomes impractical \citep{wu2026}.

Evaluation therefore increasingly relies on a sample of units or tasks \citep{polo2024, yauney2026}.
In the benchmarking literature, \citet{fogliato2024} and \citet{fisch2024} draw the questions to score at random within strata, and estimate one overall score.
\citet{wu2026} propose an adaptive sampling method under a fixed query budget, paired with a prediction from a Bayesian factor model fit on other LLMs' historical results.
For traffic monitoring, \citet{dobi2026} estimate the prevalence of policy-violating content by domain from a daily random sample.
Although benchmarking and traffic monitoring have developed separately, they both require estimating performance over a larger population using a subset of samples.

We formulate this common problem using finite-population survey sampling. 
We treat the complete evaluation set as a finite population and target the mean outcome within each reporting domain. 
This perspective also makes the goal of uncertainty quantification concrete: the true domain mean is a fixed population quantity, and a valid interval should cover it at the stated rate over repeated samples, while remaining as narrow as possible. 
Because this design-based interpretation does not require assumptions on a superpopulation distribution, it stays meaningful under distribution shift, which is common in AI evaluation as systems and their users change \citep{wu2026}.

A common approach in AI evaluation is prediction-powered inference \citep[PPI;][]{angelopoulos2023ppi}.
Under the finite-population view, PPI is equivalent to the difference estimator of classical survey sampling \citep{sarndal1992}, a correspondence first noted by \citet{fogliato2024} and formalized by \citet{mozer2026}.
Difference estimators use unit-level auxiliary information, which is data recorded on every population unit that may help predict the outcome.
PPI takes the auxiliary as a prediction of the outcome to make the estimate more precise.
PPI++ \citep{angelopoulos2023ppipp}, applied to disaggregated evaluation by \citet{emmenegger2026}, further tunes how much of the prediction to use.
Its survey sampling equivalent is the generalized regression (GREG) estimator \citep{mozer2026}.

The need to estimate many domain means from a limited sample leads naturally to small area estimation \citep{rao2015}. 
Developed for surveys with large overall samples but few observations within particular domains, small area estimation uses models to borrow strength across domains. 
Using the terminology from small area estimation, both PPI and GREG estimate a domain mean from that domain's data alone, and thus fall in a category called `direct estimators'.
For disaggregated evaluation, the sampling budget is spread across many domains.
Thus, direct estimators are imprecise for domains with few labels.

We therefore focus on so-called area-level models, which combine the domain direct estimates and their sampling variances, without requiring a model of the underlying unit-level outcomes.
The standard area-level model is the Fay--Herriot model \citep{fay1979}, which shrinks each direct estimate toward a shared regression on domain-level covariates, with greater shrinkage applied to less precise estimates.

Recent work in disaggregated AI evaluation has begun to incorporate shrinkage in estimators as well.
Empirical Bayes and hierarchical models that shrink subgroup estimates toward a shared fit have been proposed for benchmark task types and for subgroups in algorithmic fairness \citep{fogliato2024precise, herlihy2024, li2025pas}.
\citet{fogliato2024precise} note that the Fay--Herriot model is a linear special case of their approach. 
These developments suggest that shrinkage and smoothing can improve domain estimators, but raise more practical questions: which smoothing model to use and how much smoothing should a model introduce?

Choosing between model-based estimates requires formal model validation.
We can compare direct estimators based on their sampling variance, since they are typically design-unbiased, whereas a model-based estimator often introduces bias in exchange for lower variance and must be judged on its total error.
In AI evaluation, cross-validation has been used to tune an estimator \citep{herlihy2024, emmenegger2026} rather than to choose among them, and \citet{fogliato2024precise} identify model selection as an open question.
In small area estimation, the comparison is often made by an information criterion, which requires a likelihood and cannot score direct estimators.
Out-of-sample validation has only recently appeared in small area estimation \citep{kawano2026, dong2026}.
Specifically, design-based cross-validation \citep[DB-CV;][]{dong2026} is pertinent here as it allows comparisons across direct and model-based estimators.

We develop an integrated workflow for estimation and validation in disaggregated AI evaluation through the lens of small area estimation.
For estimation, we introduce prediction-powered smoothing (PP-S), a Bayesian Fay--Herriot model applied to the GREG estimate of each domain.
Its base form shrinks across all domains, and an extension (PP-TS) borrows strength along a nested hierarchy of the domains.
For validation, we propose novel refinements to DB-CV, replacing a conservative bias bound with an approximately unbiased score, for reliable comparison and error reporting.
We examine the proposed methods on one instance of each setting: a curated benchmark with verifiable grading, and deployed agent traffic graded by humans.
Outcomes are observed for every unit in both datasets, allowing us to assess every estimator and the validation score itself against the oracle.
We show that smoothing improves on the direct estimators in both point and interval estimation.
We also show that the debiased score matches selection on an independent validation sample while estimating the chosen estimator's error far more accurately.

The remainder of the paper is organized as follows.
\secref{sec:design} makes the case for a probability sample as the foundation for valid inference.
\secref{sec:methodology} sets up the problem and defines the direct estimators, the smoothing models and the validation score.
\secref{sec:results} studies them on the benchmark and on the traffic data, and \secref{sec:conclusion} concludes.

\section{The Case for Probability Sampling}\label{sec:design}

In this paper, we assume that the evaluated units are obtained from a probability sample (i.e., every unit's nonzero probability of being selected is known).
In a benchmark, the evaluator chooses which tasks to query and grade and can therefore implement such a design directly.
Yet in practice, benchmarks are often scored on a curated subset, with the questions chosen by a model fit to other LLMs' results and the full score predicted from it \citep{polo2024}.
\citet{zhang2025} find that none of these methods consistently beats the mean of a random subset once the LLM under evaluation is more accurate than those the predictor was fit on.
For ranking LLMs using a benchmark, \citet{yauney2026} find that consistently ordering two of similar accuracy takes a subset large enough that random sampling is competitive with the selection methods.
In deployed traffic, the interactions that reach a human grader are often selected by a complaint, a flag, or a reviewer's attention, and such a process is usually not a probability sample.
Without a probability design, domain estimates can be biased unless the selection process is modeled explicitly \citep{pfohl2025}.
However, the modeling assumptions on the selection process cannot be validated from the labeled data \citep{molenberghs2008}.

\begin{figure}[t]
\centering
\includegraphics[width=\textwidth]{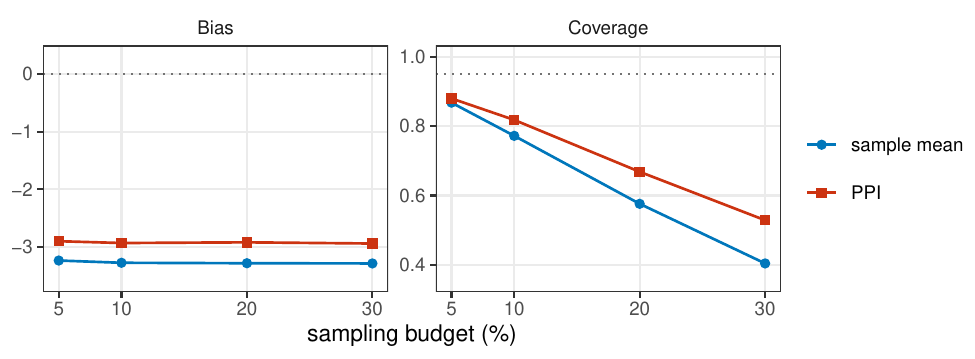}
\caption{Mean satisfaction estimated without weights from labels sampled with a bias toward rejected responses (PRISM, 63 domains, 100 replications).
The sample bias puts the correlation between being labeled and the outcome at 0.029 in magnitude at the 5\% budget.
Reported bias is in points of the 1 to 100 satisfaction scale, averaged over domains.
Coverage is of nominal 95\% intervals.
Dotted lines mark zero bias and the nominal level.}
\label{fig:hidden-selection}
\end{figure}

We illustrate the consequence of the sampling design on PRISM \citep{kirk2024}, a dataset of conversations with LLMs in which the users themselves graded every response for satisfaction.
To mimic a nonprobability sample, we draw samples in which responses that were rejected by the participant are selected at about twice the rate of the others, and then estimate as if the sample had been drawn at random (see Appendix~\ref{app:traffic-data} for the full details).
The resulting bias of the sample mean matches that of a typical online opt-in survey panel \citep{pew2023, kawano2026nps}.
Figure~\ref{fig:hidden-selection} compares the estimates with the truth.
The moderate bias persists as the sampling budget grows, while the coverage of the 95\% intervals falls well below nominal.
This is an instance of the Big Data Paradox of \citet{meng2018}: ``The bigger the data, the surer we fool ourselves.''
PPI does not fix this problem either, as without accounting for the selection process, the residuals from the labeled sample remain biased for the residuals in the full population.

Since neither the selection nor its correction can be checked from the labels alone, the rest of the paper treats a probability sample as the foundation. 
Moreover, we recommend its use for disaggregated evaluation more broadly.

\section{Methods}\label{sec:methodology}

\subsection{Direct Estimation with Auxiliary Information}\label{sec:direct}

The units making up an evaluation form a finite population $\mathcal{U} = \{1, \dots, N\}$, such as the questions of an LLM benchmark or the customer-support conversations a deployed system handled over an evaluation window.
Reporting dimensions partition $\mathcal{U}$ into $m$ domains $\mathcal{U}_1, \dots, \mathcal{U}_m$ with known sizes $N_1, \dots, N_m$.
The outcome label $y_{ij}$ is a grade of the evaluated system's performance on unit $j$ of domain $i$.
The estimand for domain $i$ is the finite-population mean
\begin{equation*} %
\theta_i \;=\; \frac{1}{N_i} \sum_{j \in \mathcal{U}_i} y_{ij}, \qquad i = 1, \dots, m.
\end{equation*}
Note that $\theta_i$ is the population mean grade or rating for a given domain $\mathcal{U}_i$.
When the given grade is binary, $\theta_i$ is the accuracy in domain $i$.

A sampling budget of $n \ll N$ labels is spent by drawing a probability sample $\mathcal{S} \subset \mathcal{U}$, in which every unit has a known probability of being labeled.
The sampled units in domain $i$ are $\mathcal{S}_i = \mathcal{S} \cap \mathcal{U}_i$, of size $n_i = \abs{\mathcal{S}_i}$, where $n = \sum_i n_i$.
Unit $j$ of domain $i$ is sampled with probability $\pi_{ij} = P(j \in \mathcal{S})$.
The problem is to estimate every $\theta_i$ from the $n$ labels, with an interval whose coverage holds over repeated draws of the sample.
The simplest estimator of $\theta_i$ weights each labeled outcome by $w_{ij} = \pi_{ij}^{-1}$,
\begin{equation*} %
\hat{\theta}_i^{\mathrm{HT}} \;=\; \frac{1}{N_i} \sum_{j \in \mathcal{S}_i} w_{ij}\, y_{ij}.
\end{equation*}
This is the Horvitz--Thompson (HT) estimator \citep{horvitzthompson1952}, which we mark with the superscript $\mathrm{HT}$ and compare the other estimators against.
Under simple random sampling (SRS) within a domain, it is the sample mean.
The HT estimator is unbiased over repeated samples.
It is also \emph{design-consistent}, i.e., it converges to the true domain mean as labels accumulate,
\begin{equation*} %
\hat{\theta}_i^{\mathrm{HT}} - \theta_i \;\xrightarrow{\;p\;}\; 0 \qquad \text{as } n_i \to \infty,
\end{equation*}
where the probability is over which sample is drawn.
\citet{dobi2026} describe an estimator of this kind deployed in real-world traffic monitoring at Pinterest.

Now suppose that every unit includes auxiliary information, recorded on all $N$ units, and write $f_{ij}$ for its value on unit $j$ of domain $i$.
Examples are the score of an LLM judge prompted to grade each interaction \citep{zheng2023}, or a question's historical pass rate among previously tested models.
Prediction-powered inference \citep[PPI;][]{angelopoulos2023ppi} treats the auxiliary as a prediction of the label,
\begin{equation}\label{eq:ppi}
\hat{\theta}_i^{\mathrm{PPI}} \;=\; \frac{1}{N_i} \sum_{j \in \mathcal{U}_i} f_{ij} \;+\; \frac{1}{N_i} \sum_{j \in \mathcal{S}_i} w_{ij}\,\{y_{ij} - f_{ij}\}.
\end{equation}
The first term is a known constant, and the second is a correction term made of weighted errors.
The weighted errors term is itself an HT estimate of the population error.
The correction term removes any systematic error in the auxiliary, so the estimator is design-consistent regardless of $f_{ij}$.
The more predictive $f_{ij}$ is of the outcome, the more variance this approach can reduce.

Rearranging \eqref{eq:ppi} allows another way to understand PPI given by 
\begin{equation*} %
\hat{\theta}_i^{\mathrm{PPI}} \;=\; \hat{\theta}_i^{\mathrm{HT}} \;+\; \left\{ \frac{1}{N_i} \sum_{j \in \mathcal{U}_i} f_{ij} \;-\; \frac{1}{N_i} \sum_{j \in \mathcal{S}_i} w_{ij}\, f_{ij} \right\},
\end{equation*}
the HT estimator plus a correction for how far the sampled units sit from the population on the auxiliary's scale.
The generalized regression (GREG) estimator extends this approach and is given by
\begin{equation*} %
\hat{\theta}_i^{\mathrm{GREG}} \;=\; \hat{\theta}_i^{\mathrm{HT}} \;+\; \hat{\lambda} \left\{ \frac{1}{N_i} \sum_{j \in \mathcal{U}_i} f_{ij} \;-\; \frac{1}{N_i} \sum_{j \in \mathcal{S}_i} w_{ij}\, f_{ij} \right\},
\end{equation*}
where $\hat{\lambda}$ is a tuning parameter which we call the tuned correction slope. 
It is the coefficient from one weighted least-squares regression of $y$ on $f$, pooled across domains.
With several unit-level covariates, the regression has one coefficient per covariate and $\hat{\lambda}$ becomes the vector of those coefficients.
Setting $\lambda = 1$ recovers $\hat{\theta}_i^{\mathrm{PPI}}$, and $\lambda = 0$ recovers the HT estimator.
GREG is also design-consistent \citep{sarndal1992}.

The power-tuned PPI \citep[PPI++;][]{angelopoulos2023ppipp} takes the same form as GREG but requires the auxiliary to be on the outcome's scale, whereas GREG fits its prediction by regression and so accepts any unit-level covariate, such as the content of a conversation in traffic.
Because of this, we refer to estimators of this type (e.g., PPI++) as GREG.
GREG's linear fit corrects the auxiliary's level and scale but not nonlinear miscalibration.
The correction slope protects GREG from a poor auxiliary, since one with no predictive value drives $\hat{\lambda}$ toward zero.

In AI evaluation, PPI has been applied with stratified sampling of the units to a single overall score on a benchmark, using an LLM judge as the prediction \citep{fisch2024} or predicted correctness based on a classifier \citep{fogliato2024}.
The prediction has also been fit to other LLMs' results on the same questions, under random sampling \citep{zhang2025} and adaptive sampling \citep{wu2026}.
\citet{emmenegger2026} apply GREG to disaggregated evaluation.

\subsection{Prediction-Powered Smoothing}\label{sec:smoothing}

PPI and GREG improve each domain's estimates using auxiliary correction, but they do not explicitly model the domain means jointly. 
Each uses only its own domain's labels, so where labels are few the estimates remain imprecise, with a variance that scales as $n_i^{-1}$.
Small area estimation addresses this imprecision with area-level models, which borrow information across domains.

The inputs into an area-level model are the design-consistent estimates $z_i$ of each $\theta_i$ and their variance $d_i$, which are typically plugged in and treated as known \citep{rao2015}.
The foundational area-level model is the Fay--Herriot model \citep{fay1979},
\begin{equation}\label{eq:fh}
\begin{aligned}
z_i \given \theta_i &\ind \N{\theta_i}{d_i}, \\
\theta_i &= \bb{x}_i^\top \bb{\beta} + v_i, \qquad v_i \iid \N{0}{\sigma^2},
\end{aligned}
\end{equation}
where $\bb{x}_i$ is a vector of domain-level covariates, $\bb{\beta}$ a coefficient vector shared by all domains, and $v_i$ a domain random effect.
The model rests on two assumptions, both of which matter most in the small domains that borrow the most.
The sampling model in \eqref{eq:fh} is a central limit approximation for the direct estimate, and it is weakest in the smallest domains.
The linking model in \eqref{eq:fh} is a modeling assumption, and misspecifying it loses precision, again in the small domains.

Conditional on $\bb{\beta}$ and $\sigma^2$, the posterior mean of $\theta_i$ is a precision-weighted compromise between the input and the regression component,
\begin{equation}\label{eq:fh-shrink}
\E[\mathrm{M}]{\theta_i \given z_i, \bb{\beta}, \sigma^2} \;=\; \gamma_i\, z_i + (1 - \gamma_i)\, \bb{x}_i^\top \bb{\beta}, \qquad \gamma_i = \frac{\sigma^2}{\sigma^2 + d_i},
\end{equation}
where $\mathbb{E}_{\mathrm{M}}$ denotes expectation under the model.
When $d_i$ is large relative to $\sigma^2$, the weight $\gamma_i$ is near zero and the estimate is close to the regression component, so a noisy domain is estimated largely by borrowing strength from the other domains, to the extent that the linking model describes them well.
When $d_i$ is small relative to $\sigma^2$, the weight is near one and a precise domain yields a model-based estimate that is close to its direct estimate.
Since $d_i$ vanishes as the domain sample size grows, the estimate converges to $z_i$ and is design-consistent whatever the linking model \citep[Section 6.1]{rao2015}.

Taking the HT estimate as the input, $z_i = \hat{\theta}_i^{\mathrm{HT}}$, gives the classical Fay--Herriot (FH) estimator.
To better incorporate auxiliary information, we take the GREG estimate as the input $z_i$, i.e.,
\begin{equation}\label{eq:pps}
\begin{aligned}
\hat{\theta}_i^{\mathrm{GREG}} \given \theta_i &\ind \N{\theta_i}{d_i}, \\
\theta_i &= \bb{x}_i^\top \bb{\beta} + v_i, \qquad v_i \iid \N{0}{\sigma^2},
\end{aligned}
\end{equation}
where $d_i$ is now the variance of the GREG estimator.
We call this approach prediction-powered smoothing (PP-S) and denote $\hat{\theta}_i^{\mathrm{PP\text{-}S}} = \E[\mathrm{M}]{\theta_i \given \hat{\bb{\theta}}^{\mathrm{GREG}}}$.
The same two-stage construction appears in survey statistics as the smoothed model-assisted estimator of \citet{gao2024}, who smooth a GREG on the logit scale for binary outcomes.

Under PP-S, the auxiliary can enter the smoother in two places and act on different terms of the weighted compromise shown in \eqref{eq:fh-shrink}.
At the unit level, it enters through the GREG estimator.
A good unit-level auxiliary reduces the variance $d_i$, so the smoothing can happen over a more precise input.
At the domain level, the auxiliary can enter through $\bb{x}_i$.
A good domain-level auxiliary improves the regression component the input is pulled toward, which matters most in the small domains.

\begin{figure}[t]
\centering
\includegraphics[width=\textwidth]{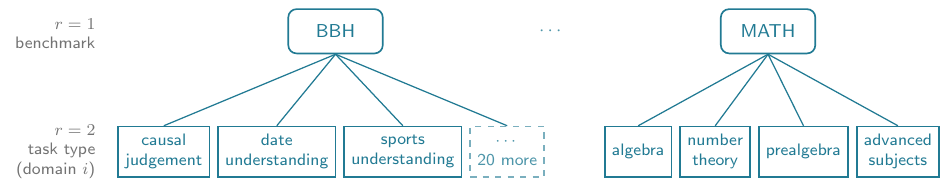}
\caption{An example taxonomy from the Open LLM Leaderboard benchmark dataset: 34 task types, the domains at level $r = 2$, nested in five benchmarks at level $r = 1$.
Two of the five benchmarks are shown.}
\label{fig:taxonomy}
\end{figure}

Under the linking model in \eqref{eq:pps}, the domain effects are modeled a priori as independent draws from a single distribution.
However, evaluation domains often provide more structure, with a domain sitting inside nested reporting categories.
We call such a nested partition of the domains a taxonomy, borrowing the term AI evaluation uses for the hierarchical category schemes its benchmarks are organized by \citep{vidgen2024, li2024salad}. 
Figure~\ref{fig:taxonomy} shows an example taxonomy from the Open LLM Leaderboard benchmark dataset.

We index the taxonomy levels by $r = 1, \dots, R$ from coarsest to finest, with level $R$ being the domains themselves.
Write $a_r(i)$ for the level-$r$ category of domain $i$, so that $a_R(i) = i$.
The single random effect can be replaced by a sum of effects per level,
\begin{equation*} %
\theta_i \;=\; \bb{x}_i^\top \bb{\beta} + \sum_{r=1}^{R} v^{(r)}_{a_r(i)}, \qquad v^{(r)}_k \iid \N{0}{\sigma_r^2},
\end{equation*}
independently across levels, with the sampling model in \eqref{eq:pps} unchanged.
We refer to this linking model as prediction-powered taxonomy smoothing (PP-TS).
Under PP-TS, domains share the effects of every category above them, so information is borrowed most strongly among the domains the taxonomy places closest together.
Each level's variance $\sigma_r^2$ is estimated from the data, so a level whose categories do not differ beyond the covariates receives a variance near zero.

When applied to the HT estimator instead of GREG, the taxonomy linking model is the Fay--Herriot model with two-fold \citep{torabi2014} or three-fold \citep{marcis2023} random effects, which we label TFH.

We fit all models using Bayesian inference, with a flat prior $p(\bb{\beta}, \sigma^2) \propto 1$ for FH and PP-S.
For TFH and PP-TS, we assign a diffuse normal prior on $\bb{\beta}$ and weakly informative inverse-gamma priors on the variances $\sigma_r^2$ to stabilize estimation.

\subsection{Design-Based Validation}\label{sec:validation}

The usefulness of the auxiliary information and smoothing is unknown in advance, so no candidate estimator is uniformly preferable. 
We therefore develop a design-based procedure for comparing the disaggregated estimators. 
We split the sampled units of each domain at random into $K$ folds so that each fold is itself a probability sample.
Write $\hat{\theta}_i^{\mathrm{HT},(k)}$ for the HT estimate from fold $k$ alone, with the weights multiplied by $K$, since the fold holds one $K$-th of the sample.
A candidate estimator $M$ fitted without fold $k$ is $\hat{\theta}_i^{M,(-k)}$.

We evaluate each candidate estimator in terms of its mean squared error under the cross-validation design,
\begin{equation}\label{eq:cv-target}
\mathrm{MSE}_i(M) \;=\; \E{\bigl(\hat{\theta}_i^{M,(-k)} - \theta_i\bigr)^2},
\end{equation}
where the expectation is over both the sampled units and the random splitting into $K$ folds.
For the choice of $K$ and its effect on validation, see \citet{mcalinn2025}.
A naive CV score estimating the target MSE can be computed by comparing the candidate estimator with the held-out HT estimate,
\begin{equation*} %
\mathrm{S}^{\mathrm{naive}}_i(M) \;=\; \frac{1}{K}\sum_{k=1}^{K}\bigl(\hat{\theta}_i^{M,(-k)} - \hat{\theta}_i^{\mathrm{HT},(k)}\bigr)^2 .
\end{equation*}

Because the score uses $\hat{\theta}_i^{\mathrm{HT},(k)}$ in place of $\theta_i$, the naive score is biased for the target error \eqref{eq:cv-target}.
The first source is the fold-level bias, which arises from the split of the sample into folds.
The held-out HT estimate for each fold is noisy and correlated with the fit on the other folds, and both shift every candidate's score.
The second source is the sample-level bias; for a given sample, every fold inherits the realized error of the HT estimator, $\hat{\theta}_i^{\mathrm{HT}} - \theta_i$, even though the HT estimator is unbiased over repeated samples.
\citet{dong2026} remove the fold-level bias, applying the correction
\begin{equation*} %
\mathrm{S}^{\mathrm{adj}}_i(M) \;=\; \mathrm{S}^{\mathrm{naive}}_i(M)
 - \frac{1}{K}\sum_{k=1}^{K}\bigl(\hat{\theta}_i^{\mathrm{HT},(k)} - \bar{\theta}_i^{\mathrm{HT}}\bigr)^2
 + \frac{2}{K}\sum_{k=1}^{K}\bigl(\hat{\theta}_i^{\mathrm{HT},(k)} - \bar{\theta}_i^{\mathrm{HT}}\bigr)\bigl(\hat{\theta}_i^{M,(-k)} - \bar{\theta}_i^{M}\bigr),
\end{equation*}
where $\bar{\theta}_i^{\mathrm{HT}}$ and $\bar{\theta}_i^{M}$ are the averages of $\hat{\theta}_i^{\mathrm{HT},(k)}$ and $\hat{\theta}_i^{M,(-k)}$ over the $K$ folds.
The two correction terms estimate the fold-level bias, with full details in Appendix~\ref{app:dbcv}.

This adjusted score does not account for the sample-level bias.
Left uncorrected, this bias favors candidates that track the HT estimate on the given sample over those that track the true domain mean (Appendix~\ref{app:adjusted}).
\citet{dong2026} propose a bound for the remaining sample-level bias instead.
Assuming that each fit without one fold centers on the full-sample fit up to a small remainder, the sample-level bias can be estimated to give a debiased score
\begin{equation}\label{eq:score-deb}
\mathrm{S}^{\mathrm{deb}}_i(M) \;=\; \mathrm{S}^{\mathrm{adj}}_i(M) - d_i + 2\,\Cov{\hat{\theta}_i^{\mathrm{HT}}}{\hat{\theta}_i^{M}},
\end{equation}
where $d_i$ here is the variance of the HT estimator, the same for every candidate $M$.
We refer to validation with this score as DB-CV.

The covariance in \eqref{eq:score-deb} is available in closed form for every candidate estimator considered here.
For the HT estimator, the covariance is the variance $d_i$.
For PPI and GREG, write $r_{ij}^{\mathrm{PPI}} = y_{ij} - f_{ij}$ and $r_{ij}^{\mathrm{GREG}} = y_{ij} - \hat{\lambda} f_{ij}$ for the residual, with $\hat{\lambda}$ held fixed as in the GREG variance.
Their covariance with the HT estimator is estimated by the bivariate form of the variance estimator that gives $d_i$,
\begin{equation*} %
\widehat{\operatorname{Cov}}\bigl(\hat{\theta}_i^{\mathrm{HT}}, \hat{\theta}_i^{\mathrm{GREG}}\bigr)
\;=\; \frac{n_i}{n_i - 1}\,
\frac{\sum_{j \in \mathcal{S}_i} w_{ij}^{2}\,(1 - \pi_{ij})\,(y_{ij} - \bar{y}_i)(r_{ij} - \bar{r}_i)}
     {\bigl(\sum_{j \in \mathcal{S}_i} w_{ij}\bigr)^{2}},
\end{equation*}
with $\bar{y}_i = \sum_{j \in \mathcal{S}_i} w_{ij}\, y_{ij} / \sum_{j \in \mathcal{S}_i} w_{ij}$ the weighted mean of the labels and $\bar{r}_i$ the same weighted mean of the residuals.

For the smoothers, the covariance is
\begin{equation*} %
\Cov{\hat{\theta}_i^{\mathrm{HT}}}{\hat{\theta}_i^{M}} \;\approx\; \frac{\Var[\mathrm{M}]{\theta_i \given \bb{z}}}{\Var{z_i}}\, \Cov{\hat{\theta}_i^{\mathrm{HT}}}{z_i},
\end{equation*}
where $\Var[\mathrm{M}]{\theta_i \given \bb{z}}$ is the posterior variance of $\theta_i$ under the model.
The approximation is exact when the variance parameters are known and holds to first order when the posterior integrates over them.
The value of $\Var{z_i}$ depends on the input direct estimator. 
When the input is the HT estimate, $\Var{z_i}$ and $\Cov{\hat{\theta}_i^{\mathrm{HT}}}{z_i}$ both equal $d_i$ and the covariance reduces to the posterior variance.
When the input is the GREG estimate, $\Var{z_i}$ is the GREG variance and $\Cov{\hat{\theta}_i^{\mathrm{HT}}}{z_i}$ is the covariance defined for GREG. 
We derive each of these in Appendix~\ref{app:cov}.

\section{Results}\label{sec:results}

We study the methods on one instance of each setting distinguished in \secref{sec:intro}.
The first is a curated benchmark with verifiable grading, the Open LLM Leaderboard as compiled by \citet{wu2026}.
Here, a sample from a benchmark is used to estimate a single LLM's performance across domains.
The second is PRISM \citep{kirk2024}, which we treat as the traffic of a deployed AI system that routes each conversation to one of many LLMs.
Each response is graded by the user who received it, and the domains are the combinations of conversation type and LLM.
We use the two studies for different purposes.
The benchmark study is a comprehensive comparison of every estimator we discuss in this paper against the oracle.
The traffic monitoring study demonstrates a workflow using a single sample and evaluates the DB-CV score.
The dataset at hand is treated as a finite population, held fixed throughout, so every $\theta_i$ is known exactly and the only randomness is which units are sampled.
Table~\ref{tab:setup} summarizes the two studies.

\begin{table}[t]
\centering
\renewcommand{\arraystretch}{1.3}
\caption{The two studies, with the population, estimand and auxiliary information of each.}
\label{tab:setup}
\begin{tabular}{>{\raggedright\arraybackslash}p{3.0cm}>{\raggedright\arraybackslash}p{5.3cm}>{\raggedright\arraybackslash}p{5.3cm}}
\toprule
 & Open LLM Leaderboard & PRISM agent traffic \\
\midrule
Setting & verifiable: graded against a known answer & open-ended: graded by a human user \\
Population & 9{,}324 questions from five benchmarks & 68{,}371 rated LLM responses from 1{,}500 participants \\
Outcome & \texttt{microsoft/phi-4}'s grade (binary) & the participant's satisfaction rating (1 to 100) \\
Domains & $m = 34$ task types nested in the 5 benchmarks & $m = 63$ crossings of 21 LLMs and 3 conversation types \\
Estimand $\theta_i$ & \texttt{phi-4}'s accuracy & mean satisfaction \\
Auxiliary information & answers from two historical models; historical difficulty & an LLM judge's score; content covariates\\
Labels at 10\% & 932, about 27 per domain & 6{,}837, about 108 per domain \\
Study scope & assessing the accuracy of every estimator & demonstrating the workflow and assessing its validation step \\
\bottomrule
\end{tabular}
\end{table}

In both studies, labels are drawn by stratified SRS without replacement, with domains as strata and allocation proportional to $N_i$.
We set the sampling budget at 10\% of each population throughout.
The appendix repeats the results at larger budgets, and similar conclusions are found.
Sampling at each budget is replicated $S = 100$ times.
The sample is a sizable share of each population, so the variance of every direct estimate includes a finite-population correction.

For each estimator, we report a point estimate $\hat{\theta}_i$ and a 95\% interval $[l_i, u_i]$ for each domain.
For direct estimators, we consider the Wald interval $\hat{\theta}_i \pm 1.96\sqrt{d_i}$ and for model-based estimators, we use the posterior credible interval.
Errors are aggregated over domains with weights $q_i$ that sum to one.
On the benchmark every task type is reported on its own, so $q_i = 1/m$.
In traffic an error matters in proportion to the traffic it affects, so $q_i = N_i/N$, the domain's traffic share.
Point estimate accuracy is summarized by the root mean squared error (RMSE),
\begin{equation*} %
\mathrm{RMSE} \;=\; \Bigl( \sum_{i=1}^{m} q_i\, (\hat{\theta}_i - \theta_i)^2 \Bigr)^{1/2}.
\end{equation*}
For the intervals we report the empirical coverage of the nominal 95\% level, the weighted fraction of domains whose interval contains $\theta_i$.
Coverage checks calibration alone, and an estimator can game it with wide intervals, so we also report the interval score \citep{gneiting2007}, a proper scoring rule that penalizes width and missed coverage together,
\begin{equation*} %
\mathrm{IS}_{\alpha} \;=\; \sum_{i=1}^{m} q_i \Bigl\{ (u_i - l_i) \;+\; \frac{2}{\alpha}\,(l_i - \theta_i)\,\mathbb{I}\{\theta_i < l_i\} \;+\; \frac{2}{\alpha}\,(\theta_i - u_i)\,\mathbb{I}\{\theta_i > u_i\} \Bigr\}, 
\end{equation*}
where $\mathbb{I}\{\cdot\}$ is the indicator function, and smaller values indicate better interval estimates.
We take $\alpha = 0.05$ in our analysis.
We report the means of these metrics across the $S$ replicate samples.

\subsection{Estimator Comparison: Open LLM Leaderboard}\label{sec:sims}

\begin{table}[t]
\centering
\small
\setlength{\tabcolsep}{5pt}
\caption{RMSE and IS of the seven estimators at the 10\% budget under each auxiliary, averaged over 100 replications (Open LLM Leaderboard, 34 domains).
Each smoother carries the auxiliary's domain mean in its linking model.
Lower is better and the best value in each column is bold.
The HT estimator uses no auxiliary, so its row is the same under all three.
Standard errors of the means are at most 0.001 for RMSE and 0.01 for IS.}
\label{tab:bench-main}
\input{tables/bench_main}
\end{table}

We use the question-by-model record compiled from the Open LLM Leaderboard by \citet{wu2026}, containing data for about 4{,}400 models.
This compilation is a valuable resource for evaluation research since the original records from the leaderboard are scattered across thousands of files.
The LLM under evaluation is \texttt{microsoft/phi-4}, and the estimand is its accuracy on each of the 34 task types.
Each task type belongs to one of five benchmarks (BBH, GPQA, IFEval, MATH and MuSR), and this two-level structure is the taxonomy shown in Figure~\ref{fig:taxonomy}.

We use three auxiliaries, all built from the leaderboard's historical records on every question, which contain neither \texttt{phi-4} nor any model fine-tuned from it.
Two of the auxiliaries are the answers of a single earlier model, scored correct or incorrect: a math specialist (\texttt{Qwen2.5-Math-7B-Instruct}) and a small generalist (\texttt{Qwen2.5-1.5B-Instruct}).
The third is historical difficulty, the share of earlier models that answered the question correctly.
The answers of the two historical models are weak auxiliaries and difficulty is a strong one, correlating 0.31, 0.33 and 0.61 with the outcome at the unit level.

The comparison covers the three direct estimators (HT, PPI and GREG) and the four smoothers: FH and TFH on the HT input, and PP-S and PP-TS on the GREG input.
Every smoother uses the auxiliary's domain mean as its only domain-level covariate.

Table~\ref{tab:bench-main} reports RMSE and IS at the 10\% budget under each auxiliary.
First we see that the prediction-powered smoothers outperform their baselines under every auxiliary on both RMSE and IS.
The gap is larger under historical difficulty, which is a strong auxiliary at both the unit and domain levels.
Notably, PP-TS outperforms all estimators, direct or smoothing, for every auxiliary.
Every estimator's coverage lies between 0.93 and 0.95 against the nominal 0.95.
The full results at every budget, including coverage, are given in Appendix~\ref{app:sims}.

The value of GREG's tuning is apparent, especially under the weak auxiliaries.
PPI has no such tuning, so under the weak auxiliaries its RMSE rises above that of HT.
PPI is designed for an auxiliary that is itself a prediction of the outcome and a weak prediction can still make it perform worse than the HT estimator.
GREG tunes the correction slope down to about a third and stays at the level of HT.
Under historical difficulty, GREG's tuned slope is close to one and the two estimators agree.

\begin{figure}[t]
\centering
\includegraphics[width=\textwidth]{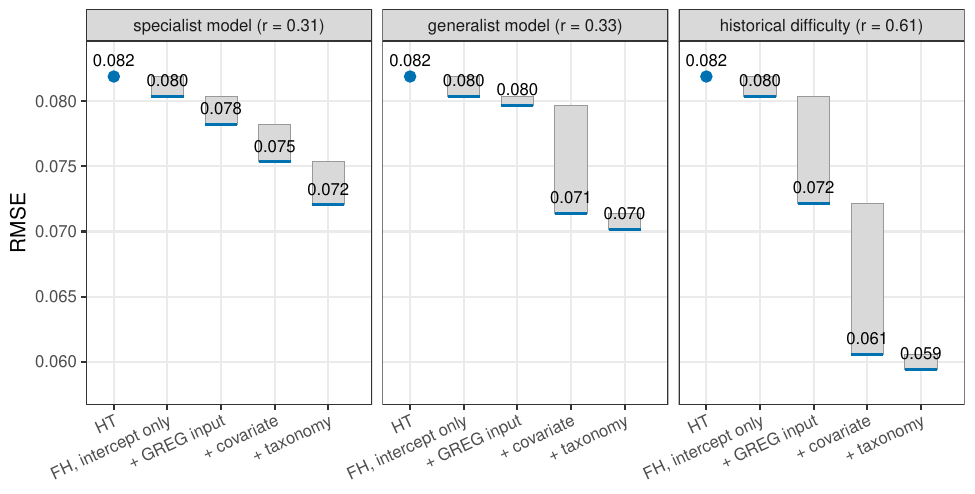}
\caption{RMSE at the 10\% budget as the components of PP-TS are added one at a time, under each auxiliary (Open LLM Leaderboard, 100 replications).
Components accumulate left to right from the HT estimator: the Fay--Herriot model on the HT input with intercept-only linking, the GREG input (PP-S with $\bb{x}_i = 1$), the auxiliary's domain mean as linking covariate (PP-S), and taxonomy smoothing (PP-TS).
The panel strips give each auxiliary's unit-level correlation $r$ with the outcome.
The printed values are entries of Table~\ref{tab:bench-main} and of the tables in Appendix~\ref{app:sims}.}
\label{fig:bench-waterfall}
\end{figure}

Figure~\ref{fig:bench-waterfall} breaks down the components of PP-TS one at a time and shows where its gain over HT comes from.
We see that pooling on its own adds little: the Fay--Herriot model with intercept-only linking on the HT input uses no auxiliary, and it barely improves on HT.
Most of the gains come from the other components.
Replacing the HT input with the GREG input helps most under the strong auxiliary, where GREG performs well.
The biggest gains generally come from the domain-level covariate. 
It is clear that an auxiliary's value at the domain level cannot be determined from its unit-level accuracy. 
The two weak auxiliaries have nearly the same unit-level correlation with the outcome, yet the generalist's domain mean lowers RMSE far more than the specialist's.
These findings likely apply to auxiliary information provided by LLM judges in general: two judges of equal accuracy can differ in what they contribute as a covariate.
Taxonomy smoothing picks up structure that the covariate leaves behind.
The specialist's domain mean errs unevenly (the specialist does well on the MATH task types but poorly on IFEval) and this pattern is exactly what taxonomy smoothing absorbs.
PP-TS therefore gains most under the specialist, and it brings the two weak auxiliaries close to even.

How much an auxiliary helps at the domain level, and how much taxonomy smoothing adds, both vary with the data at hand.
Thus, the linking structure of the smoothing model must be validated.

\subsection{Workflow and Validation: PRISM Agent Traffic}\label{sec:selection}

A real-world evaluation of a deployed AI system has three steps: design and collection of the sample, fitting candidate estimators, and validation without an oracle.
To demonstrate this, we first walk through a single-sample analysis that compares multiple candidate estimators using the DB-CV score of \secref{sec:validation}.
We then assess DB-CV against alternative validation procedures over repeated samples.

We treat the PRISM dataset as the traffic of a single deployed service that routes conversations across many LLMs and evaluate that service as a whole.
The estimand is the mean satisfaction rating in each of the 63 domains, the crossings of 21 LLMs with 3 conversation types.
The two-level structure of routed LLMs each with the 3 conversation types provides the taxonomy.
We use \texttt{gpt-5-nano} at low reasoning effort as a judge to predict each participant's own rating based on the content of each conversation (see Appendix~\ref{app:traffic-judge}).
Its score correlates 0.40 with the ratings at the unit level and 0.93 between domain means.
As before, we report the 10\% budget, which gives 6{,}837 labels or about 108 per domain.

\paragraph{Single-sample workflow.}
In our first example, the analyst is given one sample and considers seven candidate estimators: the HT estimator along with variants of GREG, PP-S and PP-TS.
GREG, PP-S, and PP-TS are each fit with the judge or with both the judge and content covariates from the conversation. 
The content covariates include the length of the response and other information about the conversation (see Appendix~\ref{app:traffic-data}).
When an auxiliary is used, it is used both in the GREG at the unit level and in the smoothing as domain means.
DB-CV uses $K = 5$ folds within strata, and its scores are aggregated over domains with the traffic-share weights $q_i = N_i/N$ and reported on the RMSE scale.

\begin{table}[h]
\centering
\caption{The candidate field on one 10\% sample (replicate 1).
The DB-CV score, on the RMSE scale, and the mean 95\% interval width are computed from the sample alone.
The last column is the candidate's oracle RMSE, weighted by traffic share, over 100 replications.
The full result across replications is given in Appendix~\ref{app:traffic}.}
\label{tab:traffic-field}
\begin{tabular}{lrrr}
\toprule
candidate & DB-CV score & interval width & oracle RMSE \\
 & & & (hidden from analyst) \\
\midrule
HT & 2.90 & 10.47 & 2.505 \\
GREG (judge) & 2.75 & 10.01 & 2.418 \\
GREG (judge + content) & 2.39 & 8.87 & 2.148 \\
PP-S (judge) & 2.43 & 8.97 & 2.253 \\
PP-S (judge + content) & 1.09 & 5.59 & 1.527 \\
PP-TS (judge) & 2.07 & 8.57 & 1.943 \\
PP-TS (judge + content) & 1.72 & 7.64 & 1.720 \\
\bottomrule
\end{tabular}
\end{table}

Table~\ref{tab:traffic-field} shows the candidate field, where every column but the last is computed from the sample alone.
DB-CV selects PP-S with the judge and content covariates, and orders all seven candidates exactly as the oracle RMSE does.
The content covariates clearly help, and through validation, the analyst can see this without the oracle. 
That is, every candidate that includes content covariates has a lower score and a narrower interval compared to its counterpart without content covariates.
With the judge alone, PP-TS beats PP-S on both the score and the oracle RMSE, but adding the content covariates reverses the order.

\begin{figure}[t]
\centering
\includegraphics[width=\textwidth]{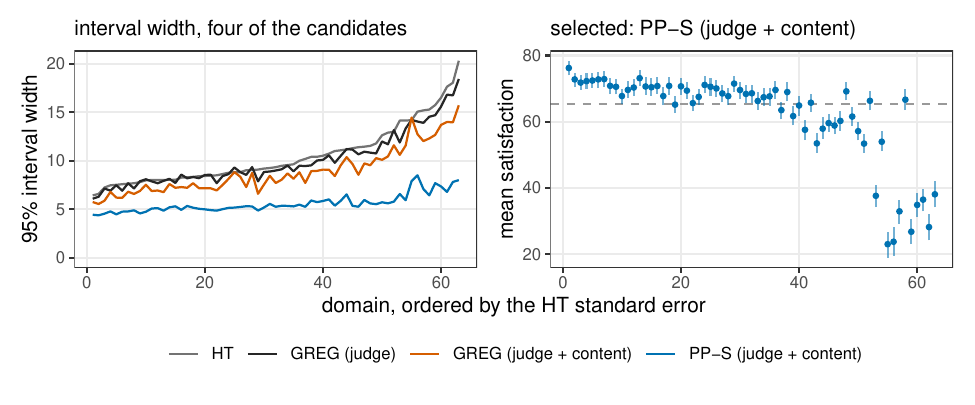}
\caption{The single-sample analysis (PRISM, 10\%, replicate 1).
Left: interval widths of four candidates over the 63 domains.
Right: the domain estimates and 95\% intervals of the selected candidate, PP-S with judge and content covariates, against the estimated traffic-wide mean (dashed).
Domains are ordered by the standard error of the HT estimator, so the right end of each panel is where smoothing has the most to do.
In the sparsest quarter of domains the intervals average 15 points for HT, 12 for GREG with all covariates and 7 for PP-S.}
\label{fig:traffic-single}
\end{figure}

Figure~\ref{fig:traffic-single} shows the interval widths of four candidates on the left and the estimates of the selected candidate on the right, with domains ordered by the standard error of the HT estimator.
Smoothing narrows the intervals most where HT is least precise, to about half of HT's width in the sparsest quarter of domains.
These sparse domains also happen to have the lowest satisfaction, so precision is gained where monitoring may be needed most.
Smoothing may add more operational value in monitoring rare outcomes, where the direct estimates can be even less precise.
\citet{dobi2026} address this with larger samples and by pooling over weeks.

\paragraph{Validation at a fixed budget.}
In the single-sample analysis, the rankings of DB-CV and the oracle RMSE agree. 
We now turn to the evaluation of the model comparison procedure over repeated samples.
We compare the proposed DB-CV procedure with three other validation procedures at the same total sampling budget over 100 replications.
The first is the naive CV score of \secref{sec:validation}, with no bias correction.
The other two split the sampling budget between two independent samples: the candidates are fit on the first and scored against the second, and the chosen candidate is then refit on the pooled labels.
We test an even 50/50 split and the 80/20 split conventional for a held-out set.
We also consider the adjusted CV score in \citet{dong2026}, but in this task, the procedure abstains on every pairwise comparison due to the conservative bounds, so its results are not reported here.

\begin{table}[t]
\centering
\caption{Four validation procedures at the same total sampling budget of 10\%, 100 replications.
Rank correlation is Kendall's $\tau$ between each procedure's scores and the oracle ordering of the seven candidates; the last column is the reported RMSE of the chosen candidate as a multiple of its oracle RMSE.
The 20\% budget is in Appendix~\ref{app:traffic}.}
\label{tab:traffic-validators}
\begin{tabular}{lrrr}
\toprule
validation procedure & RMSE of chosen & rank corr. & reported / oracle RMSE \\
\midrule
DB-CV (one sample) & 1.543 & 0.89 & 1.07$\times$ \\
naive CV (one sample) & 1.528 & 0.88 & 4.00$\times$ \\
50/50 (two samples) & 1.557 & 0.78 & 2.59$\times$ \\
80/20 (two samples) & 1.678 & 0.62 & 3.54$\times$ \\
\bottomrule
\end{tabular}
\end{table}

Table~\ref{tab:traffic-validators} grades each procedure on three dimensions. 
The first is the oracle RMSE of the candidate it chooses across samples. 
On this dimension, the two CV scores and the 50/50 split choose roughly equally well, and the 80/20 split chooses worse.
The second is the rank correlation between scores for a particular validation procedure and the oracle ordering.
Here the two CV scores order the candidates better than either split. 
The third is the error each procedure reports, relative to the oracle RMSE of its chosen candidate.
This is where DB-CV differs most from the others. 
The splits and naive CV overstate the chosen candidate's error several times over, because the held-out estimates they score against carry sampling error of their own (\secref{sec:validation}). 
Debiasing removes this, and the benefit has practical significance, as the reported error is what tells the analyst whether the chosen estimates can be trusted and whether the sampling budget was enough. 

To conclude, the proposed DB-CV score can select as well as two independent samples at the same budget and it is the only procedure whose reported error is close to the oracle. 

\section{Conclusion}\label{sec:conclusion}

This paper builds on ideas from small area estimation to develop an integrated workflow for estimation and validation in disaggregated AI evaluation.
We apply it to a benchmark with verifiable grading and to deployed agent traffic graded by humans, two settings that to our knowledge have not been treated as one estimation problem.
The proposed PP-S approach, smoothing a difference estimator across domains, improves both point and interval estimates over PPI and the other direct estimators in both settings.
Coverage of the interval estimates stays near nominal throughout.
In the benchmark study, we found that shrinkage across domains on its own accounts for little of the improvement.
The auxiliary information used at the unit and domain levels, and the reporting taxonomy, provide essential structure for the improvements from smoothing.
The structure that helps differs by dataset, so validation is a critical part of the workflow.
The debiased score we propose makes this choice from a single sample and, at equal budget, selects just as well as having two independent samples (one for training and the other for validation).
The proposed score also estimates the chosen estimator's error accurately, while the alternatives overstate it several times over.

One immediate extension concerns AI evaluation in unseen domains with no observed outcomes \citep{saxena2024predicting}.
The smoothing models may still be used to estimate performance in a domain with no labeled units whenever its domain-level covariates are available. 
Determining when such zero-shot domain evaluation is reliable remains an open question.
More broadly, there is untapped potential in the deeper integration of AI evaluation and survey methodology.
For example, a grade for an open-ended task is a judgment, and two graders may not agree on a single number.
The survey literature treats this disagreement as measurement error and carries it into the uncertainty of the estimate \citep{hansen1961}.
Another example is the large nonprobability samples that deployed systems generate, such as conversations flagged for review.
Combining such data with a probability sample guards against its selection bias \citep{kawano2026nps}.
Uncertainty quantification is essential to AI evaluation, and it is where the integration of these two fields could be most fruitful.

\section*{Data Availability}
Code reproducing every figure and table in this paper is available at \url{https://github.com/sho-kawano/disagg-ai-eval}.
The repository includes the processed Open LLM Leaderboard question-by-model tables used in the benchmark study, derived from the public release of \citet{wu2026}, and the precomputed LLM judge scores and frozen judge prompt used in the traffic study.
The PRISM dataset \citep{kirk2024} is publicly available on the Hugging Face Hub at \url{https://huggingface.co/datasets/HannahRoseKirk/prism-alignment}.

\section*{Acknowledgements}
We used Anthropic's Claude models (Opus 4.8 and 5, Fable 5 and 5.1) to assist with coding, literature search, review of derivations, LaTeX formatting, and manuscript editing.

\section*{Funding and Conflict of Interest}
None declared.

\newpage
\bibliographystyle{apalike}
\bibliography{references}

\newpage
\appendix

\section{Details for Design-Based Validation}\label{app:dbcv}

Here we provide the derivations of the debiased score of \secref{sec:validation}.
For ease of notation, we work in one domain and suppress its index.
Here $\theta$ is the domain mean and $\est^{\mathrm{HT}}$ the full-sample HT estimator, with variance $d$.
Our objective is to validate a candidate $M$, where $\est^{M}$ is its estimator computed from the full sample. 
We divide our sample into folds indexed by $k$. 
For each fold, we construct an HT estimate $\est^{\mathrm{HT},(k)}$ using the held-out fold $k$, which is used for validation.
We also construct the estimator $\est^{M,(-k)}$ based on the remaining $K-1$ folds.
We denote the averaged value of these quantities across folds as $\bar{\theta}^{\mathrm{HT}}$ and $\bar{\theta}^{M}$. 
Two sources of randomness are in play, which sample $\Samp$ is drawn and how it is split into folds. 
Every expectation below operates over both sources.

\subsection{The Bias of the Naive Score}\label{app:naive-bias}

Write
$$
e^{M,(-k)} = \est^{M,(-k)} - \theta, \qquad e^{\mathrm{HT},(k)} = \est^{\mathrm{HT},(k)} - \theta
$$
for the errors of the candidate and held-out HT estimate at fold $k$.
The target is the candidate's mean squared error averaged over the folds,
$$
\mathrm{MSE}(M) = \frac{1}{K}\sum_{k=1}^{K}\E{(e^{M,(-k)})^2}.
$$
The folds are drawn at random, so the error has the same distribution at every fold.
The naive score replaces the unknown $\theta$ by the held-out HT estimate, whose discrepancy from the candidate is the difference of the two errors,
\begin{equation*}
\mathrm{S}^{\mathrm{naive}}(M) = \frac{1}{K}\sum_{k=1}^{K}\bigl(\est^{M,(-k)} - \est^{\mathrm{HT},(k)}\bigr)^2
= \frac{1}{K}\sum_{k=1}^{K}\bigl(e^{M,(-k)} - e^{\mathrm{HT},(k)}\bigr)^2.
\end{equation*}
We first seek to show that the bias for this naive score is:
$$
\E{\mathrm{S}^{\mathrm{naive}}(M)} - \mathrm{MSE}(M) = \frac{1}{K}\sum_{k=1}^{K} \Var{\est^{\mathrm{HT},(k)}} - \frac{2}{K}\sum_{k=1}^{K}\Cov{\est^{\mathrm{HT},(k)}}{\est^{M,(-k)}}.
$$
\begin{proof}
Given the definition of the naive score as a difference of the two errors, we have
\begin{equation}\label{eq:naive_bias}
\E{\bigl(e^{M,(-k)} - e^{\mathrm{HT},(k)}\bigr)^2} = \E{(e^{M,(-k)})^2} + \E{(e^{\mathrm{HT},(k)})^2} - 2\,\E{e^{M,(-k)}\,e^{\mathrm{HT},(k)}}
\end{equation}
where this is a per-fold quantity.
Note the second term is simply the variance of the held-out HT estimate, since it is unbiased.

The last cross term can be simplified to 
\begin{align*}
\E{e^{M,(-k)}\,e^{\mathrm{HT},(k)}}
&= \Cov{e^{M,(-k)}}{e^{\mathrm{HT},(k)}} + \E{e^{M,(-k)}}\,\E{e^{\mathrm{HT},(k)}} \\
&= \Cov{\est^{M,(-k)}}{\est^{\mathrm{HT},(k)}} + \E{e^{M,(-k)}}\,\E{e^{\mathrm{HT},(k)}} \\
&= \Cov{\est^{\mathrm{HT},(k)}}{\est^{M,(-k)}} + \E{e^{M,(-k)}} \cdot 0.
\end{align*}
The second equality is due to $\theta$ being a constant and the third equality results from the unbiasedness of the held-out HT estimate.
Then \eqref{eq:naive_bias} can be simplified to
$$
\E{\bigl(e^{M,(-k)} - e^{\mathrm{HT},(k)}\bigr)^2} = \E{(e^{M,(-k)})^2} +  \Var{\est^{\mathrm{HT},(k)}} - 2\,\Cov{\est^{\mathrm{HT},(k)}}{\est^{M,(-k)}}.
$$
Thus, the expectation of the naive score is
\begin{align*}
\E{\mathrm{S}^{\mathrm{naive}}(M)} 
&= \frac{1}{K}\sum_{k=1}^{K} \E{\bigl(e^{M,(-k)} - e^{\mathrm{HT},(k)}\bigr)^2}\\
&= \frac{1}{K}\sum_{k=1}^{K} \biggl( \E{(e^{M,(-k)})^2} +  \Var{\est^{\mathrm{HT},(k)}} - 2\,\Cov{\est^{\mathrm{HT},(k)}}{\est^{M,(-k)}} \biggr)\\
&= \mathrm{MSE}(M) +  \frac{1}{K}\sum_{k=1}^{K} \Var{\est^{\mathrm{HT},(k)}} - \frac{2}{K}\sum_{k=1}^{K}\Cov{\est^{\mathrm{HT},(k)}}{\est^{M,(-k)}}.
\end{align*}
\end{proof}

Note that this bias breaks into two components when conditioned on the realized sample $\Samp$.
The variance of the held-out estimate splits as
\begin{align*}
\Var{\est^{\mathrm{HT},(k)}}
&= \E{\Var{\est^{\mathrm{HT},(k)} \given \Samp}} + \Var{\E{\est^{\mathrm{HT},(k)} \given \Samp}} \\
&= \E{\Var{\est^{\mathrm{HT},(k)} \given \Samp}} + \Var{\est^{\mathrm{HT}}} \\
&= \E{\Var{\est^{\mathrm{HT},(k)} \given \Samp}} + d.
\end{align*}
The first equality is the law of total variance.
The second holds because, given the sample, each sampled unit lands in fold $k$ with probability $1/K$, which cancels the factor $K$ on its weight, so the held-out estimate has conditional mean $\est^{\mathrm{HT}}$.\footnote{For the H\'ajek estimator, which divides by the sum of the weights rather than the domain size, this conditional mean and the fold average used in Appendix~\ref{app:adjusted} hold to first order, and exactly under simple random sampling within the domain with equal fold sizes.}
The covariance with the candidate splits in the same way,
\begin{align*}
\Cov{\est^{\mathrm{HT},(k)}}{\est^{M,(-k)}}
&= \E{\Cov{\est^{\mathrm{HT},(k)}}{\est^{M,(-k)} \given \Samp}} \\
&\quad + \Cov{\E{\est^{\mathrm{HT},(k)} \given \Samp}}{\E{\est^{M,(-k)} \given \Samp}} \\
&= \E{\Cov{\est^{\mathrm{HT},(k)}}{\est^{M,(-k)} \given \Samp}} + \Cov{\est^{\mathrm{HT}}}{\E{\est^{M,(-k)} \given \Samp}},
\end{align*}
by the law of total covariance and the same conditional mean.
Averaging the last term over the folds,
\begin{align*}
\frac{1}{K}\sum_{k=1}^{K}\Cov{\est^{\mathrm{HT}}}{\E{\est^{M,(-k)} \given \Samp}}
&= \Cov{\est^{\mathrm{HT}}}{\E{\bar{\theta}^{M} \given \Samp}} \\
&= \Cov{\est^{\mathrm{HT}}}{\bar{\theta}^{M}} - \E{\Cov{\est^{\mathrm{HT}}}{\bar{\theta}^{M} \given \Samp}} \\
&= \Cov{\est^{\mathrm{HT}}}{\bar{\theta}^{M}}.
\end{align*}
The first equality is linearity, the second is the law of total covariance for $\bar{\theta}^{M}$, and the third holds because $\est^{\mathrm{HT}}$ is fixed given the sample.
Substituting the three results into the bias of the naive score,
\begin{align}
\MoveEqLeft \E{\mathrm{S}^{\mathrm{naive}}(M)} - \mathrm{MSE}(M) \nonumber \\
&= \frac{1}{K}\sum_{k=1}^{K}\Var{\est^{\mathrm{HT},(k)}} - \frac{2}{K}\sum_{k=1}^{K}\Cov{\est^{\mathrm{HT},(k)}}{\est^{M,(-k)}} \nonumber \\
&= \frac{1}{K}\sum_{k=1}^{K}\Bigl\{\E{\Var{\est^{\mathrm{HT},(k)} \given \Samp}} + d\Bigr\} \nonumber \\
&\qquad\qquad - \frac{2}{K}\sum_{k=1}^{K}\Bigl\{\E{\Cov{\est^{\mathrm{HT},(k)}}{\est^{M,(-k)} \given \Samp}}  + \Cov{\est^{\mathrm{HT}}}{\E{\est^{M,(-k)} \given \Samp}}\Bigr\} \nonumber \\
&= \underbrace{\frac{1}{K}\sum_{k=1}^{K}\E{\Var{\est^{\mathrm{HT},(k)} \given \Samp}}  - \frac{2}{K}\sum_{k=1}^{K}\E{\Cov{\est^{\mathrm{HT},(k)}}{\est^{M,(-k)} \given \Samp}}}_{\text{fold-level bias}} \nonumber \\
&\qquad\qquad + \underbrace{d - 2\,\Cov{\est^{\mathrm{HT}}}{\bar{\theta}^{M}}}_{\text{sample-level bias}}. \label{eq:app-sample-bias}
\end{align}
In the last equality, we group the terms into fold-level and sample-level bias components.
The fold-level bias comes from the random split of a fixed sample.
The sample-level bias comes from the draw of the sample itself.
Every held-out estimate centers on the full-sample HT estimate rather than on $\theta$, so the spread across the folds carries no information about it.
We first examine the adjusted score of \citet{dong2026} that removes the fold-level bias. 

\subsection{The Adjusted Score}\label{app:adjusted}

The fold-level bias above has two terms, the average conditional variance of the held-out estimates and their average conditional covariance with the candidate.
The adjusted score of \citet{dong2026} estimates them by correcting the naive score,
$$\mathrm{S}^{\mathrm{adj}}(M) = \mathrm{S}^{\mathrm{naive}}(M) - \hat{\upsilon} + 2\,\hat{c}^{M},$$
where $\hat{\upsilon}$ and $\hat{c}^{M}$ are defined by
\begin{equation*}
\hat{\upsilon} = \frac{1}{K}\sum_{k=1}^{K}\bigl(\est^{\mathrm{HT},(k)} - \bar{\theta}^{\mathrm{HT}}\bigr)^2, \qquad
\hat{c}^{M} = \frac{1}{K}\sum_{k=1}^{K}\bigl(\est^{\mathrm{HT},(k)} - \bar{\theta}^{\mathrm{HT}}\bigr)\bigl(\est^{M,(-k)} - \bar{\theta}^{M}\bigr).
\end{equation*}
We seek to show that
\begin{equation*}
\E{\hat{\upsilon}} = \frac{1}{K}\sum_{k=1}^{K}\E{\Var{\est^{\mathrm{HT},(k)} \given \Samp}}, \qquad
\E{\hat{c}^{M}} = \frac{1}{K}\sum_{k=1}^{K}\E{\Cov{\est^{\mathrm{HT},(k)}}{\est^{M,(-k)} \given \Samp}}.
\end{equation*}

\begin{proof}
The proof uses two facts about the held-out HT estimates, which follow from how the folds are built.
Write $\Samp^{(k)}$ for the sampled units in fold $k$, so that $\est^{\mathrm{HT},(k)} = N^{-1}\sum_{j \in \Samp^{(k)}} K\,w_j\,y_j$.
The first fact is that the held-out estimates average to the full-sample estimate,
\begin{equation}\label{eq:app-fold-average}
\bar{\theta}^{\mathrm{HT}}
= \frac{1}{K}\sum_{k=1}^{K}\frac{1}{N}\sum_{j \in \Samp^{(k)}} K\,w_j\,y_j
= \frac{1}{N}\sum_{j \in \Samp} w_j\,y_j
= \est^{\mathrm{HT}},
\end{equation}
where the second equality cancels the factor $K$ and uses that each sampled unit sits in exactly one fold.
The second fact is that, given the sample, each held-out estimate has conditional mean equal to the full-sample estimate,
\begin{equation}\label{eq:app-cond-mean}
\E{\est^{\mathrm{HT},(k)} \given \Samp}
= \frac{1}{N}\sum_{j \in \Samp} K\,w_j\,y_j\,P\bigl(j \in \Samp^{(k)} \given \Samp\bigr)
= \frac{1}{N}\sum_{j \in \Samp} w_j\,y_j
= \est^{\mathrm{HT}},
\end{equation}
where the first equality takes the expectation over the split with the weights and labels fixed, and the second uses that each sampled unit lands in fold $k$ with probability $1/K$.

Given the sample, the expectation of $\hat{\upsilon}$ is
\begin{align*}
\E{\hat{\upsilon} \given \Samp}
&= \frac{1}{K}\sum_{k=1}^{K}\E{\bigl(\est^{\mathrm{HT},(k)} - \bar{\theta}^{\mathrm{HT}}\bigr)^2 \given \Samp} \\
&= \frac{1}{K}\sum_{k=1}^{K}\E{\bigl(\est^{\mathrm{HT},(k)} - \est^{\mathrm{HT}}\bigr)^2 \given \Samp} \\
&= \frac{1}{K}\sum_{k=1}^{K}\Var{\est^{\mathrm{HT},(k)} \given \Samp}.
\end{align*}
The first equality is linearity of conditional expectation, and the second uses \eqref{eq:app-fold-average}.
The third uses \eqref{eq:app-cond-mean}, since a squared deviation from the conditional mean has conditional expectation equal to the conditional variance.

The expectation of $\hat{c}^{M}$ takes two steps.
First, the candidate's fold average $\bar{\theta}^{M}$ drops out,
\begin{align*}
\E{\hat{c}^{M} \given \Samp}
&= \E{\frac{1}{K}\sum_{k=1}^{K}\bigl(\est^{\mathrm{HT},(k)} - \est^{\mathrm{HT}}\bigr)\bigl(\est^{M,(-k)} - \bar{\theta}^{M}\bigr) \given \Samp} \\
&= \mathbb{E}\biggl[\,\frac{1}{K}\sum_{k=1}^{K}\bigl(\est^{\mathrm{HT},(k)} - \est^{\mathrm{HT}}\bigr)\est^{M,(-k)} - \bar{\theta}^{M}\,\frac{1}{K}\sum_{k=1}^{K}\bigl(\est^{\mathrm{HT},(k)} - \est^{\mathrm{HT}}\bigr) \Bigm| \Samp\,\biggr] \\
&= \frac{1}{K}\sum_{k=1}^{K}\E{\bigl(\est^{\mathrm{HT},(k)} - \est^{\mathrm{HT}}\bigr)\est^{M,(-k)} \given \Samp}.
\end{align*}
The first equality uses \eqref{eq:app-fold-average}, and the second expands the product and takes $\bar{\theta}^{M}$, which does not depend on $k$, outside the second sum.
The third holds because the second sum is $\bar{\theta}^{\mathrm{HT}} - \est^{\mathrm{HT}}$, which is zero by \eqref{eq:app-fold-average}.

Second, each summand is a conditional covariance.
Write $\mu^{(k)} = \E{\est^{M,(-k)} \given \Samp}$ for the conditional mean of the candidate's fit, which is fixed given the sample.
Then
\begin{align*}
\E{\bigl(\est^{\mathrm{HT},(k)} - \est^{\mathrm{HT}}\bigr)\est^{M,(-k)} \given \Samp}
&= \E{\bigl(\est^{\mathrm{HT},(k)} - \est^{\mathrm{HT}}\bigr)\bigl(\est^{M,(-k)} - \mu^{(k)}\bigr) \given \Samp} \\
&= \Cov{\est^{\mathrm{HT},(k)}}{\est^{M,(-k)} \given \Samp}.
\end{align*}
The first equality subtracts $\mu^{(k)}$, which changes nothing because $\mu^{(k)}$ is fixed given the sample and $\est^{\mathrm{HT},(k)} - \est^{\mathrm{HT}}$ has conditional mean zero by \eqref{eq:app-cond-mean}.
The second holds because, by \eqref{eq:app-cond-mean}, both factors are deviations from their conditional means.

Finally, we average the two conditional expectations over samples,
\begin{align*}
\E{\hat{\upsilon}}
&= \E{\E{\hat{\upsilon} \given \Samp}}
= \frac{1}{K}\sum_{k=1}^{K}\E{\Var{\est^{\mathrm{HT},(k)} \given \Samp}}, \\
\E{\hat{c}^{M}}
&= \E{\E{\hat{c}^{M} \given \Samp}}
= \frac{1}{K}\sum_{k=1}^{K}\E{\Cov{\est^{\mathrm{HT},(k)}}{\est^{M,(-k)} \given \Samp}}.
\end{align*}
In each line, the first equality is the law of total expectation, and the second substitutes the conditional expectation computed above.
\end{proof}

The adjusted score is therefore left with the sample-level bias,
\begin{align*}
\E{\mathrm{S}^{\mathrm{adj}}(M)} - \mathrm{MSE}(M)
&= d - 2\,\Cov{\est^{\mathrm{HT}}}{\bar{\theta}^{M}} \\
&= d\,(1 - 2\rho),
\end{align*}
where $\rho = \Cov{\est^{\mathrm{HT}}}{\bar{\theta}^{M}} / d$.
The bias is zero only at $\rho = 1/2$.
The HT estimator has $\rho = 1$, so its adjusted score is too low by $d$.
A candidate that uses no labels has $\rho = 0$, so its adjusted score is too high by $d$.

\subsection{The Debiased Score}\label{app:debias}

The sample-level bias is $d - 2\,\Cov{\est^{\mathrm{HT}}}{\bar{\theta}^{M}}$, which involves the average of the candidate's $K$ training fits, while the debiased score of \secref{sec:validation} uses the full-sample fit $\est^{M}$.
We assume that, given the sample, each training fit centers on the full-sample fit up to a remainder of order $n^{-1}$, with $n$ the size of the whole sample,
\begin{equation}\label{eq:app-cand-mean}
\E{\est^{M,(-k)} \given \Samp} = \est^{M} + O_p\bigl(n^{-1}\bigr).
\end{equation}
\citet{dong2026} derive this approximation for the difference of two candidates from a linearization condition, and we assume it for a single candidate.
Under this assumption,
\begin{align*}
d - 2\,\Cov{\est^{\mathrm{HT}}}{\bar{\theta}^{M}}
&= d - 2\,\Cov{\est^{\mathrm{HT}}}{\E{\bar{\theta}^{M} \given \Samp}} \\
&= d - 2\,\Cov{\est^{\mathrm{HT}}}{\est^{M} + O_p\bigl(n^{-1}\bigr)} \\
&\approx d - 2\,\Cov{\est^{\mathrm{HT}}}{\est^{M}}.
\end{align*}
The first equality holds because $\est^{\mathrm{HT}}$ is fixed given the sample, and the second averages \eqref{eq:app-cand-mean} over the folds.
Subtracting this sample-level bias from the adjusted score gives the debiased score \eqref{eq:score-deb},
\begin{equation*}
\mathrm{S}^{\mathrm{deb}}(M) = \mathrm{S}^{\mathrm{adj}}(M) - d + 2\,\Cov{\est^{\mathrm{HT}}}{\est^{M}}.
\end{equation*}

\subsection{The Covariance for Each Candidate}\label{app:cov}

The debiased score needs $d$ and the covariance $\Cov{\est^{\mathrm{HT}}}{\est^{M}}$, and only the covariance depends on the candidate.
Both are computed from the full sample.

\paragraph{HT.}
When the candidate is the HT estimator itself, the covariance is its variance $d$, so the debiased score adds $-d + 2d = d$ to the adjusted score.

\paragraph{PPI and GREG.}
Both estimators are a known constant plus a weighted mean of residuals,
\begin{equation*}
\est^{M} = \frac{1}{N}\sum_{j \in \mathcal{U}} p_j + \frac{1}{N}\sum_{j \in \Samp} w_j\,r_j,
\end{equation*}
where $p_j$ is the prediction, equal to $f_j$ under PPI and $\hat{\lambda} f_j$ under GREG, and $r_j = y_j - p_j$ is the residual.
With $\hat{\lambda}$ treated as fixed, the first term is a constant, so
\begin{align*}
\Cov{\est^{\mathrm{HT}}}{\est^{M}}
&= \Cov{\frac{1}{N}\sum_{j \in \Samp} w_j\,y_j}{\frac{1}{N}\sum_{j \in \mathcal{U}} p_j + \frac{1}{N}\sum_{j \in \Samp} w_j\,r_j} \\
&= \Cov{\frac{1}{N}\sum_{j \in \Samp} w_j\,y_j}{\frac{1}{N}\sum_{j \in \Samp} w_j\,r_j}.
\end{align*}
The first equality writes out both estimators, and the second drops the constant.
The result is the covariance between the weighted means of the labels and of the residuals.
It is not the candidate's own variance, which would have the residuals in place of the labels in the first argument.

Both $d$ and this covariance are estimated from the full sample by the linearization estimator for a weighted mean,
\begin{align*}
\hat{d} &= \frac{n}{n-1}\,\frac{\sum_{j \in \Samp} w_j^2\,(1 - \pi_j)\,(y_j - \bar{y})^2}{\bigl(\sum_{j \in \Samp} w_j\bigr)^2}, \\
\widehat{\operatorname{Cov}}\bigl(\est^{\mathrm{HT}}, \est^{M}\bigr) &= \frac{n}{n-1}\,\frac{\sum_{j \in \Samp} w_j^2\,(1 - \pi_j)\,(y_j - \bar{y})(r_j - \bar{r})}{\bigl(\sum_{j \in \Samp} w_j\bigr)^2},
\end{align*}
where $n$ is the domain's sample size, $\bar{y} = \sum_{j \in \Samp} w_j\,y_j / \sum_{j \in \Samp} w_j$ is the weighted mean of the labels, and $\bar{r} = \sum_{j \in \Samp} w_j\,r_j / \sum_{j \in \Samp} w_j$ is the weighted mean of the residuals.
The covariance estimator replaces one of the two centered labels in $\hat{d}$ by a centered residual.
The factor $1 - \pi_j$ is the finite-population correction, and $n/(n-1)$ is the small-sample factor.

\paragraph{Smoothers.}
A smoother uses every domain, so we bring back the domain index and write $z_i$ for the input of domain $i$.
Once $\bb{\beta}$ and the variance parameters are fixed ($\sigma^2$ under FH and PP-S, and $\sigma_r^2$ under TFH and PP-TS), each of the four smoothers is $\bb{A}\bb{z}$ plus a term that does not depend on $\bb{z}$, for a matrix $\bb{A}$ that we treat as fixed.
Then
\begin{align*}
\Cov{\est_i^{\mathrm{HT}}}{\est_i^{M}}
&= \Cov{\est_i^{\mathrm{HT}}}{\sum_{l=1}^{m} A_{il}\,z_l} \\
&= \sum_{l=1}^{m} A_{il}\,\Cov{\est_i^{\mathrm{HT}}}{z_l} \\
&= A_{ii}\,\Cov{\est_i^{\mathrm{HT}}}{z_i}.
\end{align*}
The first equality writes out row $i$ of $\bb{A}\bb{z}$ and drops the constant term, and the second is bilinearity with $\bb{A}$ fixed.
The third holds because sampling errors of the direct estimates are independent across domains.

The coefficient $A_{ii}$ has a closed form.
Stacking the domains, the four smoothers of \secref{sec:smoothing} share the form
\begin{equation*}
\bb{z} \given \bb{\theta} \sim \N{\bb{\theta}}{\bb{D}}, \qquad
\bb{\theta} = \bb{X}\bb{\beta} + \bb{v}, \qquad
\bb{v} \sim \N{\bb{0}}{\bb{G}},
\end{equation*}
where $\bb{D} = \operatorname{diag}\{\Var{z_1}, \dots, \Var{z_m}\}$ holds the sampling variances of the inputs, $\bb{X}$ stacks the covariates, and $\bb{G}$ is the covariance of the random effects.
Given $\bb{\beta}$, the sampling error $\bb{z} - \bb{\theta}$ is independent of $\bb{\theta}$, so $\Cov[\mathrm{M}]{\bb{\theta}}{\bb{z}} = \bb{G}$ and $\Var[\mathrm{M}]{\bb{z}} = \bb{G} + \bb{D}$, and conditioning on $\bb{z}$ gives
\begin{align*}
\bb{A} &= \bb{G}(\bb{G} + \bb{D})^{-1}, \\
\Var[\mathrm{M}]{\bb{\theta} \given \bb{z}}
&= \bb{G} - \bb{G}(\bb{G} + \bb{D})^{-1}\bb{G} \\
&= \bb{G}(\bb{G} + \bb{D})^{-1}(\bb{G} + \bb{D}) - \bb{G}(\bb{G} + \bb{D})^{-1}\bb{G} \\
&= \bb{G}(\bb{G} + \bb{D})^{-1}\bigl(\bb{G} + \bb{D} - \bb{G}\bigr) \\
&= \bb{A}\bb{D}.
\end{align*}
Because $\bb{D}$ is diagonal, the $i$-th diagonal entry gives $A_{ii} = \Var[\mathrm{M}]{\theta_i \given \bb{z}} / \Var{z_i}$, so
\begin{equation*}
\Cov{\est_i^{\mathrm{HT}}}{\est_i^{M}} = \Var[\mathrm{M}]{\theta_i \given \bb{z}}\,\frac{\Cov{\est_i^{\mathrm{HT}}}{z_i}}{\Var{z_i}}.
\end{equation*}
Under FH and TFH the input is the HT estimate, so the ratio is one and the covariance is the posterior variance.
Under PP-S and PP-TS the input is the GREG estimate.
The numerator of the ratio is then the GREG covariance above, and the denominator is the GREG variance already supplied to the smoother.

The derivation above holds the variance parameters fixed.
The implementation instead fits each smoother by Markov chain Monte Carlo (MCMC), which integrates over them, so the smoothed estimate is no longer of the form $\bb{A}\bb{z}$ plus a constant.
A first-order expansion of $\est_i^{M}$ in the inputs takes the place of that linear form, and the covariance chain above becomes
\begin{align*}
\Cov{\est_i^{\mathrm{HT}}}{\est_i^{M}}
&\approx \sum_{l=1}^{m} \frac{\partial \est_i^{M}}{\partial z_l}\,\Cov{\est_i^{\mathrm{HT}}}{z_l} \\
&= \frac{\partial \est_i^{M}}{\partial z_i}\,\Cov{\est_i^{\mathrm{HT}}}{z_i}.
\end{align*}
This is the earlier chain with $A_{il}$ replaced by $\partial \est_i^{M} / \partial z_l$, and the equality again uses the independence of the sampling errors across domains.
With the variance parameters fixed, $\partial \est_i^{M} / \partial z_i = A_{ii}$ and the two chains coincide.

To find the derivative, note that $\est_i^{M} = \E[\mathrm{M}]{\theta_i \given \bb{z}}$ is a posterior mean, and apply Tweedie's formula \citep{efron2011}.
Under the normal sampling model and any prior on $\bb{\theta}$, including one that integrates over the variance parameters, the formula and its companion for the posterior variance are
\begin{equation*}
\est_i^{M} = z_i + \Var{z_i}\,\frac{\partial \ell(\bb{z})}{\partial z_i}, \qquad
\Var[\mathrm{M}]{\theta_i \given \bb{z}} = \Var{z_i} + \Var{z_i}^2\,\frac{\partial^2 \ell(\bb{z})}{\partial z_i^2},
\end{equation*}
where $\ell(\bb{z})$ is the log density of the inputs under the model.
The first follows from differentiating $\ell$ once in $z_i$, which brings down the posterior mean of $\theta_i - z_i$.
Differentiating a second time brings down its posterior second moment, and subtracting the square of the first derivative leaves the posterior variance, which gives the second.
Differentiating the first,
\begin{align*}
\frac{\partial \est_i^{M}}{\partial z_i}
&= 1 + \Var{z_i}\,\frac{\partial^2 \ell(\bb{z})}{\partial z_i^2} \\
&= \frac{1}{\Var{z_i}}\biggl(\Var{z_i} + \Var{z_i}^2\,\frac{\partial^2 \ell(\bb{z})}{\partial z_i^2}\biggr) \\
&= \frac{\Var[\mathrm{M}]{\theta_i \given \bb{z}}}{\Var{z_i}},
\end{align*}
where the second equality factors out $1/\Var{z_i}$, and the third recognizes the term in parentheses as the posterior variance from the companion formula.
Substituting into the first-order chain gives
\begin{equation*}
\Cov{\est_i^{\mathrm{HT}}}{\est_i^{M}} \approx \frac{\Var[\mathrm{M}]{\theta_i \given \bb{z}}}{\Var{z_i}} \, \Cov{\est_i^{\mathrm{HT}}}{z_i},
\end{equation*}
the same formula as in the fixed case, with the posterior variance now computed from the retained draws.

\clearpage
\section{Additional Details for the Benchmark Simulation}\label{app:sims}

We merge the MATH task types on advanced subjects into one domain and exclude one task type on which \texttt{phi-4} answers every question correctly, leaving 34 task types with $N_i$ from 146 to 670, median 250.
Labels are drawn by SRS without replacement within each domain, so $\pi_{ij} = n_i / N_i$ and the variance of the HT estimator is estimated by
\begin{equation*} %
d_i \;=\; \Bigl(1 - \frac{n_i}{N_i}\Bigr) \frac{s_i^2}{n_i},
\end{equation*}
with $s_i^2$ the sample variance of the labels in $\Samp_i$.
PPI and GREG use the same expression on their residuals.
Every smoother is fit by Gibbs sampling with one chain, keeping 2{,}000 draws after a burn-in of 1{,}000 iterations.
FH and PP-S are fit on the original scale under the flat prior on $\bb{\beta}$ and $\sigma^2$.
For TFH and PP-TS the inputs $z_i$ are standardized by the mean and standard deviation of the observed inputs, with $d_i$ scaled to match, and each coefficient in $\bb{\beta}$ then has a $\N{0}{10^4}$ prior and each level variance $\sigma_r^2$ an inverse-gamma prior with shape 3 and scale 0.6.
The point estimate is the posterior mean, and the 95\% interval runs from the 2.5\% to the 97.5\% posterior quantile.

The compilation of \citet{wu2026} sorts about 4{,}400 models by release date and takes the earlier half, about 2{,}200 models, as historical.
All three auxiliaries are built from that half alone, and \texttt{phi-4} and every model fine-tuned from it fall in the newer half.

The same 100 samples serve every estimator and every auxiliary, so the rows of the estimators that use no auxiliary repeat exactly across the three auxiliary blocks.
A domain whose sampled labels all agree has an estimated variance of zero and is left out of every estimator's metrics in that replication; this affects 0.6 domains on average across the 100 replications at 10\%, 0.1 at 20\% and none at 30\%.
Tables~\ref{tab:bench-rmse}, \ref{tab:bench-is} and \ref{tab:bench-coverage} report RMSE, interval score and coverage at all three sampling budgets and under all three auxiliaries, with each smoother twice, once with intercept-only linking and once with the auxiliary's domain mean as covariate.
At a matched linking model the smoother on the GREG input is never worse than the same smoother on the HT input, under any auxiliary at any budget and paired within replication.

\input{tables/bench_rmse}
\clearpage
\input{tables/bench_is}
\input{tables/bench_coverage}

\clearpage
\section{Additional Details for the Agent Traffic Analysis}

\subsection{The PRISM Data}\label{app:traffic-data}

We use the PRISM dataset \citep{kirk2024}, which records human ratings on every response to an LLM prompt.
The dataset is large enough to sample from, and we treat it as the traffic of a deployed AI system.
In PRISM, a participant's opening prompt is sent to up to four LLMs, and the participant rates every response and chooses one LLM to continue the conversation with.
Each later turn returns two responses from the chosen LLM, which the participant again rates before choosing one.
Every rated response is a unit, so one conversation contributes several units to the population.
The content covariates are recorded for every response, whether or not its rating is sampled:
\begin{itemize}
\item the log of the response's length in characters,
\item an indicator that the response answers the opening prompt,
\item an indicator that the participant chose the response.
\end{itemize}
The traffic analysis also uses the score of an LLM judge, described in the next subsection.

In the traffic analysis of the Results, ratings are drawn by stratified SRS, so the chance that a response is sampled depends only on its domain.
Whether the participant chose a response therefore plays no part in which ratings are sampled, and the chosen indicator enters only as a covariate.

The demonstration in \secref{sec:design} instead uses informative selection, in which the chance that a response is sampled depends on something related to its rating.
Participants rate the responses they choose far higher, 81.7 on average against 54.1 for the rest.
Each response is sampled independently with probability proportional to $\exp(t z)$, where $z$ is the standardized chosen indicator and $t = -0.30$, scaled so that the expected sample size meets the budget.
A chosen response is therefore sampled at 0.54 times the rate of a rejected one, which puts the correlation between being sampled and the rating at about 0.03 in magnitude at the 5\% budget.
The judge is nearly blind to this choice.
The judge's scores differ by about 5 points between chosen and rejected responses, while the participants' ratings differ by about 28.

\subsection{The Judge}\label{app:traffic-judge}

The traffic judge is \texttt{gpt-5-nano} at low reasoning effort, prompted zero-shot to predict the participant's own 1 to 100 rating of the final response.
Each conversation is submitted in full, with the response to score marked and no participant attributes included, and the judge returns a score and a confidence as structured output.
The prompt was written once and frozen before any scoring:

\begin{quote}
\ttfamily\raggedright
You will see a conversation between a user and an AI assistant. The conversation ends with an assistant response marked for scoring. Immediately after that response, the user rated it on a scale from 1 to 100, where higher means they were more satisfied with it. Predict the score the user gave. Judge the response from that user's perspective in the context of the conversation, not against an external standard of quality.

Reply with JSON: \\
\char`\{"score": <integer 1-100>, "confidence": <integer 0-100>\char`\} \\
where confidence is how sure you are of the prediction.
\end{quote}

Scoring all 68{,}371 responses cost about six dollars.

\subsection{Results at the 20\% Budget}\label{app:traffic}

Tables~\ref{tab:traffic-truth-app} and \ref{tab:traffic-validators-app} repeat the traffic study's comparisons at the 20\% budget, 13{,}674 ratings, alongside the 10\% columns of the main text.
Domain sizes $N_i$ run from 423 to 1{,}809, median 1{,}129.
Every conclusion carries over: the oracle ordering of the candidates is unchanged, the two one-sample scores and the even split remain equivalent on the choice, and only DB-CV reports a calibrated error.
Table~\ref{tab:traffic-truth-app} also reports each candidate's interval score and coverage against the oracle.
Coverage sits between 0.943 and 0.964 against the nominal 95\% level for every candidate, and the interval-score gains track the RMSE gains.

\input{tables/traffic_truth}

\begin{table}[H]
\centering
\caption{The four validation procedures at both budgets, 100 replications, columns as in the main text.}
\label{tab:traffic-validators-app}
\begin{tabular}{llrrr}
\toprule
budget & validation procedure & RMSE of chosen & rank corr. & reported / oracle RMSE \\
\midrule
10\% & DB-CV (one sample) & 1.543 & 0.89 & 1.07$\times$ \\
     & naive CV (one sample) & 1.528 & 0.88 & 4.00$\times$ \\
     & 50/50 (two samples) & 1.557 & 0.78 & 2.59$\times$ \\
     & 80/20 (two samples) & 1.678 & 0.62 & 3.54$\times$ \\
\midrule
20\% & DB-CV (one sample) & 1.209 & 0.82 & 1.11$\times$ \\
     & naive CV (one sample) & 1.190 & 0.84 & 3.67$\times$ \\
     & 50/50 (two samples) & 1.232 & 0.74 & 2.37$\times$ \\
     & 80/20 (two samples) & 1.247 & 0.60 & 3.46$\times$ \\
\bottomrule
\end{tabular}
\end{table}

\end{document}

%% file: macros.tex
\newcommand{\abs}[1]{\lvert #1 \rvert}
\newcommand{\bb}[1]{\boldsymbol{#1}}

\newcommand{\est}{\hat{\theta}}
\newcommand{\Samp}{\mathcal{S}}
\newcommand{\iid}{\stackrel{\text{iid}}{\sim}}
\newcommand{\ind}{\stackrel{\text{ind}}{\sim}}
\newcommand{\given}{\,\mid\,}
\providecommand{\E}{}\renewcommand{\E}[2][]{\mathbb{E}_{#1}\!\left[\,#2\,\right]}
\providecommand{\Var}{}\renewcommand{\Var}[2][]{\operatorname{Var}_{#1}\!\left[\,#2\,\right]}
\providecommand{\Cov}{}\renewcommand{\Cov}[3][]{\operatorname{Cov}_{#1}\!\left[\,#2,\,#3\,\right]}

\providecommand{\N}{}\renewcommand{\N}[2]{N\!\left(#1, #2\right)}

\newcommand{\secref}[1]{Section~\ref{#1}}

%% file: tables/bench_main.tex
\begin{tabular}{lcccccc}
\toprule
& \multicolumn{2}{c}{Specialist model} & \multicolumn{2}{c}{Generalist model} & \multicolumn{2}{c}{Historical difficulty} \\
\cmidrule(lr){2-3} \cmidrule(lr){4-5} \cmidrule(lr){6-7}
Estimator & RMSE & IS & RMSE & IS & RMSE & IS \\
\midrule
$\hat{\theta}_i^{\mathrm{HT}}$ & 0.082 & 0.418 & 0.082 & 0.418 & 0.082 & 0.418 \\
$\hat{\theta}_i^{\mathrm{PPI}}$ & 0.106 & 0.498 & 0.105 & 0.503 & 0.074 & 0.358 \\
$\hat{\theta}_i^{\mathrm{GREG}}$ & 0.080 & 0.393 & 0.082 & 0.409 & 0.074 & 0.358 \\
\midrule
$\hat{\theta}_i^{\mathrm{FH}}$ & 0.078 & 0.377 & 0.072 & 0.362 & 0.067 & 0.335 \\
$\hat{\theta}_i^{\mathrm{PP\text{-}S}}$ & 0.075 & 0.357 & 0.071 & 0.355 & 0.061 & 0.301 \\
\midrule
$\hat{\theta}_i^{\mathrm{TFH}}$ & 0.074 & 0.366 & 0.071 & 0.359 & 0.065 & 0.326 \\
$\hat{\theta}_i^{\mathrm{PP\text{-}TS}}$ & \textbf{0.072} & \textbf{0.350} & \textbf{0.070} & \textbf{0.351} & \textbf{0.059} & \textbf{0.295} \\
\bottomrule
\end{tabular}

%% file: tables/bench_rmse.tex
\begin{table}[h]
\centering
\small
\setlength{\tabcolsep}{4pt}
\caption{RMSE over the 34 domains, averaged over 100 replications, at each sampling budget and under each auxiliary. Lower is better and the best value in each column is bold.}
\label{tab:bench-rmse}
\begin{tabular}{lccccccccc}
\toprule
& \multicolumn{3}{c}{Specialist model} & \multicolumn{3}{c}{Generalist model} & \multicolumn{3}{c}{Historical difficulty} \\
\cmidrule(lr){2-4} \cmidrule(lr){5-7} \cmidrule(lr){8-10}
Estimator & 10\% & 20\% & 30\% & 10\% & 20\% & 30\% & 10\% & 20\% & 30\% \\
\midrule
$\hat{\theta}_i^{\mathrm{HT}}$ & 0.0819 & 0.0555 & 0.0408 & 0.0819 & 0.0555 & 0.0408 & 0.0819 & 0.0555 & 0.0408 \\
$\hat{\theta}_i^{\mathrm{PPI}}$ & 0.1056 & 0.0700 & 0.0525 & 0.1047 & 0.0702 & 0.0526 & 0.0736 & 0.0496 & 0.0359 \\
$\hat{\theta}_i^{\mathrm{GREG}}$ & 0.0800 & 0.0541 & 0.0398 & 0.0816 & 0.0550 & 0.0406 & 0.0739 & 0.0498 & 0.0360 \\
\midrule
$\hat{\theta}_i^{\mathrm{FH}}$, intercept & 0.0804 & 0.0550 & 0.0405 & 0.0804 & 0.0550 & 0.0405 & 0.0804 & 0.0550 & 0.0405 \\
$\hat{\theta}_i^{\mathrm{FH}}$, covariate & 0.0776 & 0.0539 & 0.0400 & 0.0724 & 0.0522 & 0.0392 & 0.0675 & 0.0497 & 0.0382 \\
$\hat{\theta}_i^{\mathrm{PP\text{-}S}}$, intercept & 0.0782 & 0.0537 & 0.0395 & 0.0797 & 0.0545 & 0.0402 & 0.0722 & 0.0493 & 0.0357 \\
$\hat{\theta}_i^{\mathrm{PP\text{-}S}}$, covariate & 0.0753 & 0.0526 & 0.0390 & 0.0714 & 0.0517 & 0.0389 & 0.0606 & 0.0448 & 0.0336 \\
\midrule
$\hat{\theta}_i^{\mathrm{TFH}}$, intercept & 0.0781 & 0.0539 & 0.0402 & 0.0781 & 0.0539 & 0.0402 & 0.0781 & 0.0539 & 0.0402 \\
$\hat{\theta}_i^{\mathrm{TFH}}$, covariate & 0.0737 & 0.0521 & 0.0393 & 0.0707 & 0.0512 & 0.0388 & 0.0652 & 0.0485 & 0.0377 \\
$\hat{\theta}_i^{\mathrm{PP\text{-}TS}}$, intercept & 0.0768 & 0.0531 & 0.0394 & 0.0780 & 0.0537 & 0.0399 & 0.0712 & 0.0488 & 0.0356 \\
$\hat{\theta}_i^{\mathrm{PP\text{-}TS}}$, covariate & \textbf{0.0720} & \textbf{0.0512} & \textbf{0.0384} & \textbf{0.0702} & \textbf{0.0511} & \textbf{0.0385} & \textbf{0.0594} & \textbf{0.0441} & \textbf{0.0334} \\
\bottomrule
\end{tabular}
\end{table}

%% file: tables/bench_is.tex
\begin{table}[h]
\centering
\small
\setlength{\tabcolsep}{4pt}
\caption{Interval score of the nominal 95\% intervals, averaged over 100 replications, at each sampling budget and under each auxiliary. Lower is better and the best value in each column is bold.}
\label{tab:bench-is}
\begin{tabular}{lccccccccc}
\toprule
& \multicolumn{3}{c}{Specialist model} & \multicolumn{3}{c}{Generalist model} & \multicolumn{3}{c}{Historical difficulty} \\
\cmidrule(lr){2-4} \cmidrule(lr){5-7} \cmidrule(lr){8-10}
Estimator & 10\% & 20\% & 30\% & 10\% & 20\% & 30\% & 10\% & 20\% & 30\% \\
\midrule
$\hat{\theta}_i^{\mathrm{HT}}$ & 0.4181 & 0.2723 & 0.1974 & 0.4181 & 0.2723 & 0.1974 & 0.4181 & 0.2723 & 0.1974 \\
$\hat{\theta}_i^{\mathrm{PPI}}$ & 0.4982 & 0.3244 & 0.2441 & 0.5035 & 0.3287 & 0.2476 & 0.3583 & 0.2362 & 0.1685 \\
$\hat{\theta}_i^{\mathrm{GREG}}$ & 0.3926 & 0.2593 & 0.1887 & 0.4093 & 0.2643 & 0.1939 & 0.3580 & 0.2363 & 0.1681 \\
\midrule
$\hat{\theta}_i^{\mathrm{FH}}$, intercept & 0.3860 & 0.2632 & 0.1926 & 0.3860 & 0.2632 & 0.1926 & 0.3860 & 0.2632 & 0.1926 \\
$\hat{\theta}_i^{\mathrm{FH}}$, covariate & 0.3766 & 0.2607 & 0.1911 & 0.3622 & 0.2502 & 0.1881 & 0.3354 & 0.2377 & 0.1824 \\
$\hat{\theta}_i^{\mathrm{PP\text{-}S}}$, intercept & 0.3650 & 0.2506 & 0.1836 & 0.3805 & 0.2550 & 0.1897 & 0.3404 & 0.2296 & 0.1653 \\
$\hat{\theta}_i^{\mathrm{PP\text{-}S}}$, covariate & 0.3567 & 0.2493 & 0.1823 & 0.3551 & 0.2444 & 0.1846 & 0.3008 & 0.2129 & 0.1588 \\
\midrule
$\hat{\theta}_i^{\mathrm{TFH}}$, intercept & 0.3805 & 0.2561 & 0.1882 & 0.3805 & 0.2561 & 0.1882 & 0.3805 & 0.2561 & 0.1882 \\
$\hat{\theta}_i^{\mathrm{TFH}}$, covariate & 0.3661 & 0.2503 & 0.1856 & 0.3591 & 0.2454 & 0.1852 & 0.3257 & 0.2317 & 0.1792 \\
$\hat{\theta}_i^{\mathrm{PP\text{-}TS}}$, intercept & 0.3642 & 0.2470 & 0.1820 & 0.3770 & 0.2503 & 0.1863 & 0.3388 & 0.2250 & 0.1628 \\
$\hat{\theta}_i^{\mathrm{PP\text{-}TS}}$, covariate & \textbf{0.3497} & \textbf{0.2415} & \textbf{0.1777} & \textbf{0.3515} & \textbf{0.2404} & \textbf{0.1817} & \textbf{0.2954} & \textbf{0.2085} & \textbf{0.1566} \\
\bottomrule
\end{tabular}
\end{table}

%% file: tables/bench_coverage.tex
\begin{table}[h]
\centering
\small
\setlength{\tabcolsep}{4pt}
\caption{Empirical coverage of the nominal 95\% intervals, averaged over 100 replications, at each sampling budget and under each auxiliary. The value closest to nominal in each column is bold.}
\label{tab:bench-coverage}
\begin{tabular}{lccccccccc}
\toprule
& \multicolumn{3}{c}{Specialist model} & \multicolumn{3}{c}{Generalist model} & \multicolumn{3}{c}{Historical difficulty} \\
\cmidrule(lr){2-4} \cmidrule(lr){5-7} \cmidrule(lr){8-10}
Estimator & 10\% & 20\% & 30\% & 10\% & 20\% & 30\% & 10\% & 20\% & 30\% \\
\midrule
$\hat{\theta}_i^{\mathrm{HT}}$ & 0.935 & 0.941 & 0.952 & 0.935 & 0.941 & 0.952 & 0.935 & 0.941 & 0.952 \\
$\hat{\theta}_i^{\mathrm{PPI}}$ & 0.937 & \textbf{0.947} & \textbf{0.951} & 0.939 & 0.941 & 0.948 & 0.943 & 0.943 & 0.951 \\
$\hat{\theta}_i^{\mathrm{GREG}}$ & 0.941 & 0.942 & 0.946 & 0.941 & 0.945 & \textbf{0.950} & 0.942 & 0.942 & 0.953 \\
\midrule
$\hat{\theta}_i^{\mathrm{FH}}$, intercept & 0.941 & 0.937 & 0.951 & 0.941 & 0.937 & 0.951 & 0.941 & 0.937 & 0.951 \\
$\hat{\theta}_i^{\mathrm{FH}}$, covariate & 0.942 & 0.938 & 0.951 & 0.940 & 0.940 & 0.951 & 0.941 & 0.940 & \textbf{0.950} \\
$\hat{\theta}_i^{\mathrm{PP\text{-}S}}$, intercept & \textbf{0.946} & 0.942 & 0.946 & 0.942 & \textbf{0.947} & 0.951 & 0.943 & 0.945 & 0.954 \\
$\hat{\theta}_i^{\mathrm{PP\text{-}S}}$, covariate & 0.945 & 0.943 & 0.948 & \textbf{0.943} & 0.945 & 0.952 & 0.944 & \textbf{0.945} & 0.955 \\
\midrule
$\hat{\theta}_i^{\mathrm{TFH}}$, intercept & 0.932 & 0.933 & 0.949 & 0.932 & 0.933 & 0.949 & 0.932 & 0.933 & 0.949 \\
$\hat{\theta}_i^{\mathrm{TFH}}$, covariate & 0.934 & 0.935 & 0.949 & 0.938 & 0.936 & 0.951 & \textbf{0.946} & 0.937 & 0.951 \\
$\hat{\theta}_i^{\mathrm{PP\text{-}TS}}$, intercept & 0.932 & 0.938 & 0.944 & 0.933 & 0.939 & 0.948 & 0.931 & 0.942 & 0.952 \\
$\hat{\theta}_i^{\mathrm{PP\text{-}TS}}$, covariate & 0.936 & 0.940 & 0.949 & 0.940 & 0.942 & 0.949 & 0.942 & 0.943 & 0.954 \\
\bottomrule
\end{tabular}
\end{table}

%% file: tables/traffic_truth.tex
\begin{table}[H]
\centering
\small
\setlength{\tabcolsep}{4pt}
\caption{The candidate field against the oracle at both budgets, weighted by traffic share over 100 replications: RMSE, the interval score, and the empirical coverage of the nominal 95\% intervals. HT is the Horvitz--Thompson estimator. The best RMSE, the best interval score, and the coverage closest to nominal in each budget are bold.}
\label{tab:traffic-truth-app}
\begin{tabular}{lrrrrrr}
\toprule
 & \multicolumn{3}{c}{10\%} & \multicolumn{3}{c}{20\%} \\
\cmidrule(lr){2-4} \cmidrule(lr){5-7}
candidate & RMSE & IS & coverage & RMSE & IS & coverage \\
\midrule
HT & 2.505 & 11.45 & \textbf{0.950} & 1.684 & 7.71 & 0.948 \\
GREG (judge) & 2.418 & 11.03 & 0.950 & 1.615 & 7.41 & \textbf{0.949} \\
GREG (judge + content) & 2.148 & 9.93 & 0.949 & 1.451 & 6.57 & 0.951 \\
PP-S (judge) & 2.253 & 10.23 & 0.945 & 1.563 & 7.13 & 0.946 \\
PP-S (judge + content) & \textbf{1.527} & \textbf{7.03} & 0.943 & \textbf{1.190} & \textbf{5.43} & 0.945 \\
PP-TS (judge) & 1.943 & 9.01 & 0.964 & 1.408 & 6.53 & 0.957 \\
PP-TS (judge + content) & 1.720 & 8.10 & 0.964 & 1.267 & 5.83 & 0.959 \\
\bottomrule
\end{tabular}
\end{table}